\documentclass[a4paper,fleqn]{cas-sc}

\usepackage[authoryear]{natbib}
\usepackage{amsmath,amssymb}
\usepackage{booktabs}
\usepackage{multirow}
\usepackage{placeins}
\usepackage{float}
\usepackage{microtype}

\newif\ifanonymous
\ifdefined\ANONYMOUS
  \anonymoustrue
\else
  \anonymousfalse
\fi

\graphicspath{{figures/}{figs/}{../figs/}}

\makeatletter
\AtBeginDocument{\RenewDocumentCommand\printorcid{}{}}
\makeatother

\begin{document}
\let\WriteBookmarks\relax

\shorttitle{Auditing Architectural Complexity for Ultrasound Classification}
\ifanonymous
\shortauthors{Anonymous authors}
\else
\shortauthors{Y. Song et~al.}
\fi

\title[mode=title]{A Controlled Audit of Architectural Complexity in
Uncertainty-Aware Multi-Organ Ultrasound Classification}

\ifanonymous
\author[1]{Anonymous Author(s)}
\affiliation[1]{organization={Author names and affiliations are listed in the
separate title page},
country={}}
\else
\author[1]{Yang Song}
\cormark[1]
\fnmark[1]
\ead{yangss.song@connect.polyu.hk}
\credit{Conceptualization, Methodology, Software, Validation, Formal analysis,
Writing -- original draft, Writing -- review \& editing}

\affiliation[1]{organization={Department of Mechanical Engineering,
The Hong Kong Polytechnic University},
city={Kowloon},
country={Hong Kong}}

\author[2,3]{Pengbo Sun}
\fnmark[1]
\credit{Methodology, Software, Validation, Data curation,
Writing -- original draft}
\affiliation[2]{organization={School of Electronic and Information Engineering,
Beihang University},
city={Beijing},
postcode={100191},
country={China}}
\affiliation[3]{organization={Hangzhou International Innovation Institute,
Beihang University},
city={Hangzhou},
postcode={311115},
country={China}}

\author[4]{Shichang Feng}
\credit{Investigation, Validation, Visualization}
\affiliation[4]{organization={College of Artificial Intelligence,
Nanjing University of Aeronautics and Astronautics},
city={Nanjing},
country={China}}

\author[5]{Ye Zhu}
\credit{Methodology, Resources, Writing -- review \& editing}
\affiliation[5]{organization={Independent Researcher},
city={Jinan, Shandong},
country={China}}

\author[1]{Xin Xu}
\credit{Software, Data curation}

\author[6]{Ziran Wang}
\cormark[2]
\ead{wangziran@sdu.edu.cn}
\credit{Supervision, Project administration, Funding acquisition,
Writing -- review \& editing}
\affiliation[6]{organization={Key Laboratory of High Efficiency and Clean
Mechanical Manufacture, Ministry of Education, School of Mechanical Engineering,
Shandong University},
postcode={250000},
city={Jinan},
country={China}}

\cortext[cor1]{Corresponding author}
\cortext[cor2]{Corresponding author}
\fntext[fn1]{These authors contributed equally to this work.}
\fi

\begin{abstract}
Multi-organ ultrasound classifiers increasingly combine attention,
mixture-of-experts routing, uncertainty gating, and evidential deep learning
(EDL) objectives to address heterogeneous anatomy and acquisition. Yet a plausible design rationale
does not by itself establish that an added component improves the trained
system. Our artificial-intelligence contribution is a controlled
complexity-audit framework. Its engineering application is the deployment
decision between the maximal evidential candidate Full-EDL and simpler
alternatives. Six candidates were evaluated on the primary
dataset and three in an internal replication. Both experiments used ten matched
seeds, frozen image-level partitions, capacity- and optimisation-aware
comparisons, symmetric temperature scaling, paired decision rules, and a
separate out-of-distribution (OOD) veto. Retaining
Full-EDL did not establish a reliable macro-F1 gain on either dataset, while
the simplified alternatives remained inconclusive under the non-inferiority
margin. Simple cross-entropy with temperature scaling (Simple-CE+TS) met the
calibrated negative log-likelihood criterion on both datasets and showed
favourable descriptive selective-risk ordering. The raw calibration advantage of evidential
training on the primary dataset disappeared after temperature scaling and did
not recur on the second dataset. The gate had negligible observable influence
at the audited checkpoints, and deleting the Full-only chain revealed no stable task or
calibrated-loss benefit. Simple-CE nevertheless triggered the OOD veto against
the fetal probe but not the lung probe, precluding an unconditional OOD-safety
claim. We therefore selected Simple-CE+TS for
the evaluated in-distribution objective while retaining Full-EDL as the maximal
reference. These findings show that components should earn retention through
functional and retraining-based evidence. Calibration and distribution-shift
reliability should be evaluated separately.
\end{abstract}


\maketitle

\section{Introduction}\label{sec:intro}
Ultrasound provides portable, real-time imaging without ionising radiation, so
it is well suited to point-of-care and resource-constrained applications
\citep{Shaddock2022PortablePOCUS}. Extending automated analysis across organs
could improve the reuse of compact imaging and computing infrastructure. Doing
so, however, exposes a model to substantial variation in anatomy, acquisition
source, image quality, and class composition \citep{Jiao2024USFM}. In this setting, a useful intelligent system must be
assessed on more than its mean classification score. Its probability estimates
should support calibrated interpretation, its confidence ordering should permit
selective referral, and its behaviour under distribution shift should be
evaluated explicitly \citep{Kurz2022UncertaintyMedical,Ovadia2019Trust}. These
requirements must also be balanced against structural complexity. Every
additional routing, uncertainty, or fallback component adds to the burden of
implementation, verification, and deployment.

Several modelling ideas provide reasonable motivations for increasing system
complexity. Spatial attention can reweight local features
\citep{Woo2018CBAM}, while mixture-of-experts architectures route inputs
through specialised computational paths to accommodate heterogeneous patterns
\citep{Jacobs1991MoE,Riquelme2021VMoE}. Medical-imaging variants have also used
expert routing to address variation across domains, modalities, or lesion
types \citep{Jiang2024MedMoE,Chen2025COME}. A shared fallback path and a
sample-level reliability gate can provide an alternative to the routed expert
output. Evidential deep learning (EDL) offers a single-forward-pass
representation of class evidence and Dirichlet uncertainty
\citep{Sensoy2018EDL}. Together,
spatial reliability weighting, routed experts, shared features, and uncertainty
modulation form a plausible maximal design for multi-organ ultrasound
classification. Their individual motivations, however, do not establish that
the resulting components contribute complementary or measurable system-level
benefits when trained jointly.

The unresolved question is therefore not whether another motivated component
can be added, but whether each layer of complexity earns retention under a
controlled system-level comparison. A change in mean performance alone can be
difficult to interpret when candidates differ in optimisation, effective
capacity, checkpoint selection, or stochastic training variation.
Probabilistic objectives add a second source of ambiguity. Raw ECE
characterises native confidence, whereas evaluation after matched post-hoc
calibration asks how well each output can be calibrated using
held-out data \citep{Guo2017Calibration}. Macro-F1, calibrated loss,
selective-risk ordering, and OOD separation also evaluate distinct behaviours.
They need not favour the same candidate
\citep{Ovadia2019Trust,Geifman2018AURC}. A
defensible retention decision therefore requires several forms of evidence:
functional tests, capacity- and optimisation-aware retraining, matched
stochastic comparisons, symmetric calibration, and endpoint-specific acceptance
rules. No single favourable mean or reliability label is sufficient.

\begin{figure}[pos=!htb]
  \centering
  \includegraphics[width=\linewidth]{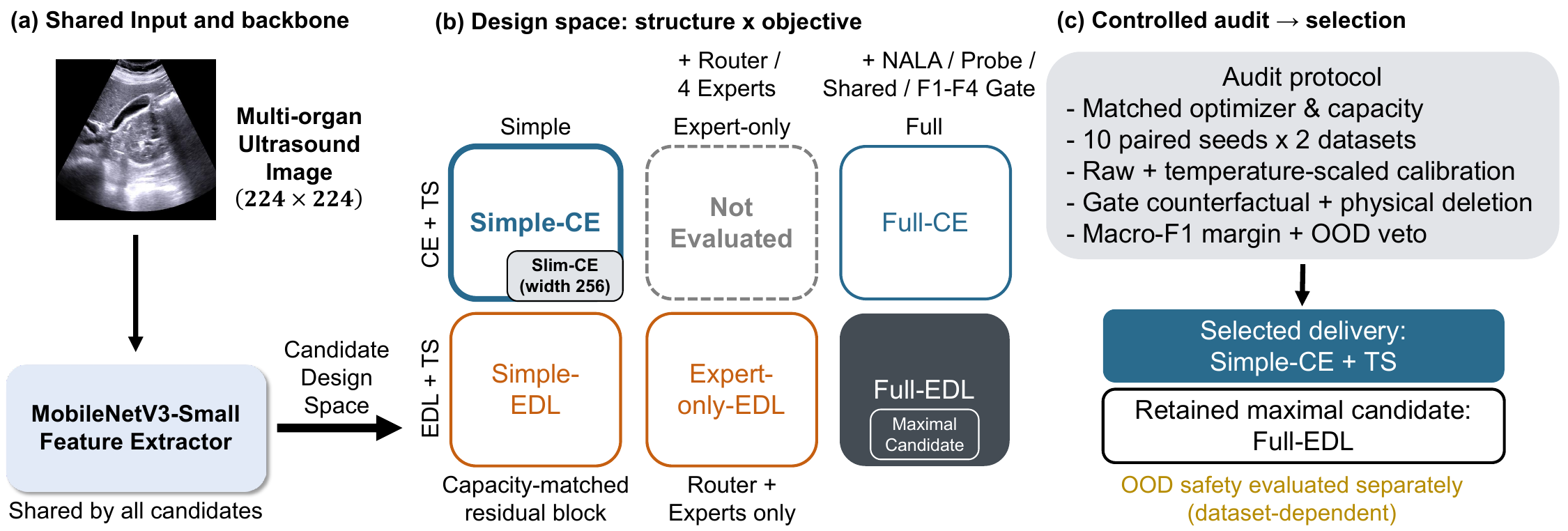}
  \caption{Controlled candidate hierarchy and engineering-selection logic.
  (a) All candidates share an ImageNet-pretrained MobileNetV3-Small feature
  extractor. (b) The evaluated design space separates structural complexity
  (Simple, Expert-only, and Full) from the training objective (CE or EDL).
  Expert-only-CE was not evaluated, and Slim-CE is a secondary lower-capacity
  operating point. (c) The audit combines matched optimisation, capacity-aware
  comparison, ten paired seeds, symmetric calibration, fixed-checkpoint gate
  counterfactuals, physical module deletion, and endpoint-specific decisions.
  Simple-CE+TS is selected for the in-distribution objective, whereas
  Full-EDL is retained as the maximal reference and OOD safety is evaluated
  separately.}
  \label{fig:hierarchy}
\end{figure}

We address this gap by asking which elements of a motivated maximal design earn
engineering retention. Full-EDL combines spatial reliability weighting, an
auxiliary evidential probe, sparse expert routing, shared fallback, and
parameter-free uncertainty modulation. It serves as the audited reference, not
as a presumed optimum. The evaluation begins with fixed-checkpoint functional
counterfactuals and then moves to physical module deletion and capacity-matched
retraining. As illustrated in Fig.~\ref{fig:hierarchy}, it covers the
available Simple, Expert-only, and Full structures under cross-entropy (CE) or
EDL objectives.
Dataset~2 provides a
six-candidate primary audit, and Dataset~1 provides a three-candidate internal
replication. Each uses ten matched seeds, frozen image-level partitions,
identity-verified optimiser grouping, and a dedicated calibration subset.
Paired rules evaluate macro-F1 and TS-NLL relative to Full-EDL. A separate OOD
veto and restricted zero-shot breast analysis define the reliability and
transfer boundaries. The design allows either greater complexity or greater
simplicity to prevail on the evidence.

The contributions of this study are fourfold. First, we develop and apply a
multilevel complexity-audit framework. It combines fixed-checkpoint functional
counterfactuals, physical module deletion, capacity-matched retraining,
structure/objective comparison, symmetric post-hoc calibration, and paired
multi-seed decision rules. Second, across two multi-organ ultrasound
collections, we show that retaining the Full-EDL component chain did not
establish a reliable macro-F1 gain. The evaluated gate behaviour also had
negligible observable influence at the audited checkpoints. Third, we
demonstrate that symmetric temperature scaling changed the apparent CE--EDL
calibration ranking. The raw-ECE advantage observed for EDL on Dataset~2 did
not recur on Dataset~1, while Simple-CE met the TS-NLL criterion on both
datasets. Fourth, we combine task performance, calibrated loss, selective
prediction, OOD vetoes, and a restricted cross-source breast stress test. This
evidence supports Simple-CE+TS for the in-distribution operating objective and
exposes probe-dependent OOD limits.

\section{Related work}\label{sec:related}

\subsection{Compact and multi-organ ultrasound intelligence}

Deep learning has supported ultrasound classification and analysis across
fetal, breast, and other anatomical applications. Recent systems have improved
task performance through stacked ensembles, multiscale or graph-based feature
integration, and ultrasound-specific representation learning
\citep{krishna2024standard,wang2025abus,Kang2024UltrasoundMIM}. Universal
ultrasound models have also investigated transfer across tasks and organs
\citep{Jiao2024USFM}. Compact backbones such as MobileNetV3 provide a
practical foundation for resource-aware implementations
\citep{Howard2019MobileNetV3}. Lightweight ultrasound classifiers have also
used plug-and-play multiscale and cross-modal fusion modules
\citep{zhu2025lightweight}. Recent applied-AI studies have explored
segmentation-prior ultrasound classification, class-imbalance-aware medical
transfer learning, and data-efficient knowledge distillation for compact
medical classifiers
\citep{jiang2025prior,jindal2025class,el2024hdkd}. These
directions primarily address predictive performance, transferable
representation, or computational design. We ask a complementary engineering
question. Once a compact backbone is augmented with multiple reliability and
routing components, which additions retain measurable value under matched
optimisation, capacity, calibration, and stochastic evaluation? The focus
shifts from proposing another high-scoring architecture to testing whether the
added complexity is justified for the intended operating objective.

\subsection{Mixture-of-experts and adaptive routing}
Mixture-of-experts models combine multiple computational paths through
input-dependent routing, with motivations ranging from local specialisation to
conditional computation \citep{Jacobs1991MoE,Shazeer2017OutrageouslyMoE}.
Sparse expert layers have been extended to vision \citep{Riquelme2021VMoE}
and, more recently, to medical systems addressing heterogeneous domains,
modalities, and tasks \citep{Jiang2024MedMoE,Li2025MoESAM,Chen2025COME}.
Related expert-system research has also explored MoE structures outside medical
imaging \citep{wang2025mixture}. These studies establish the architectural
plausibility of adaptive routing. They do not, however, quantify how strongly a
router, multiple experts, or a reliability gate influences the outputs of a
trained model. Functional counterfactuals test realised forward behaviour.
Physically deleted and retrained variants ask whether the corresponding
computational graph merits retention. We combine these views to audit one
compact uncertainty-aware MoE configuration, without generalising the result to
all expert-routing systems.

\subsection{Evidential uncertainty and post-hoc calibration}
Evidential deep learning maps non-negative class evidence to Dirichlet
parameters, providing class probabilities and an evidence-strength-based
uncertainty quantity in a single forward pass \citep{Sensoy2018EDL}. This
formulation has motivated medical applications in uncertainty-aware diagnosis,
segmentation, and OOD detection
\citep{Hung2024EvidentialProstate,Xia2024ClassBalancedEDL,Fu2025DEDL}, alongside
broader work on risk-aware classification \citep{csensoy2025risk}. Post-hoc
temperature scaling addresses a different question: it rescales a trained
model's logits using held-out calibration data without changing the predicted
class \citep{Guo2017Calibration}. Native and temperature-scaled outputs should
therefore be reported together when comparing EDL with CE. Applying calibration
to only one objective would make the comparison asymmetric. The need to assess
and calibrate predictive uncertainty has also been demonstrated in applied
engineering settings \citep{yang2024towards}. Multiple endpoints are needed
because NLL, binned ECE, Brier score, selective-risk AURC, and OOD AUROC
characterise different aspects of probabilistic behaviour
\citep{Kurz2022UncertaintyMedical}. OOD conclusions additionally depend on the
chosen score and evaluation set rather than on model confidence alone
\citep{cofta2025predictability}. This design separates native confidence properties
from post-hoc calibratability and from distribution-shift detection.

\subsection{Ablation and evidence-based model selection}
Component ablation remains an important way to assess whether an architectural
addition affects a trained system. An engineering-retention decision, however,
requires several further distinctions. A module can be present without
materially altering forward behaviour. A smaller graph can also differ from a
capacity-reduced model, and failure to demonstrate an advantage is not
equivalent to demonstrating non-inferiority. Repeated matched seeds help
characterise stochastic training variation \citep{Lemay2022MCDropoutMedical},
while prespecified margins make the tolerated performance loss and the meaning
of an inconclusive comparison explicit. For applied expert systems, these
statistical distinctions must be combined with implementation and
operating-context considerations. The highest observed mean is not a
sufficient basis for selection \citep{saibene2021expert}. Our framework extends ordinary ablation through
fixed-checkpoint intervention, physical deletion, capacity-aware retraining, and
endpoint-specific decision rules. Macro-F1, calibrated loss, selective-risk
ordering, OOD vetoes, and structural complexity remain separate. This avoids an
opaque composite reliability score.

\section{Materials and methods}\label{sec:method}

\subsection{Problem formulation and engineering objective}
\label{sec:problem}
We considered ten-class multi-organ ultrasound image classification. Given an
image $\mathbf{x}$, each candidate model $f_{\theta}$ produced a logit vector
$\mathbf{z}=f_{\theta}(\mathbf{x})\in\mathbb{R}^{10}$. Raw class probabilities
were $\mathbf{p}=\operatorname{softmax}(\mathbf{z})$, and the predicted class
was $\hat{y}=\arg\max_k p_k$. After checkpoint selection, temperature-scaled
probabilities were defined as
$\mathbf{p}^{(T)}=\operatorname{softmax}(\mathbf{z}/T)$, where $T$ was fitted on
the dedicated calibration partition as described in Section~\ref{sec:metrics}.
CE- and EDL-trained candidates therefore shared the same ten-class prediction
and evaluation space, although their training objectives differed. Maximum
softmax probability provided the confidence score for selective prediction. The
OOD score conventions are defined separately in Section~\ref{sec:metrics}.

The engineering objective was to determine whether the added components of a
motivated maximal design justified their retention. We evaluated task
performance, post-hoc probabilistic calibration, selective-risk ordering, and
OOD separation. Structural complexity and computational cost were considered
alongside these endpoints. Full-EDL served as the maximal reference candidate,
not as a presumed optimum. The macro-F1, TS-NLL, and OOD rules constrained the
interpretation of each comparison. We did not reduce the endpoints to a
single score or universal architecture ranking. Figure~\ref{fig:hierarchy}
summarises the shared backbone, candidate design space, controlled audit, and
delivery-selection process.

\subsection{Candidate design hierarchy}\label{sec:candidates}

All audit candidates used an ImageNet-pretrained MobileNetV3-Small feature
extractor and were organised along separate structural and training-objective
axes. Simple physically omitted NALA, the evidential probe, router, experts,
shared fallback, and reliability gate, while using one residual bottleneck
before pooling and classification. Expert-only retained the evidential router,
four experts, and the load-balancing term, while physically omitting NALA, the
auxiliary probe, shared fallback, and the F1--F4 reliability gate. Full
retained NALA spatial-reliability weighting, the auxiliary evidential probe,
the evidential router, four experts, shared fallback, and the parameter-free
F1--F4 reliability gate. The objective axis distinguished class-weighted cross-entropy (CE) from
evidential deep learning (EDL). Scalar temperature scaling was applied
symmetrically after training and was therefore neither an architectural
component nor a separate training objective. Table~\ref{tab:candidates}
summarises the candidate definitions.

\begin{table}[pos=!htb]
\centering
\caption{Candidate design matrix. All candidates use the same
MobileNetV3-Small feature extractor. Temperature scaling is applied
symmetrically after training and is not an architectural component.}
\label{tab:candidates}
\footnotesize
\renewcommand{\arraystretch}{1.08}
\setlength{\tabcolsep}{3pt}
\resizebox{\linewidth}{!}{%
\begin{tabular}{@{}lccccccccc@{}}
\toprule
\textbf{Candidate} & \textbf{Width} & \textbf{NALA} & \textbf{Probe} &
\textbf{Router+4 experts} & \textbf{Shared+gate} & \textbf{Objective} &
\textbf{Params (M)} & \textbf{MACs (M)} & \textbf{Evaluated on} \\
\midrule
Slim-CE            & 256  & No  & No  & No  & No  & CE  & 1.229 & 69.4  & D2 \\
Simple-CE          & 1280 & No  & No  & No  & No  & CE  & 2.411 & 127.2 & D1, D2 \\
Simple-EDL         & 1280 & No  & No  & No  & No  & EDL & 2.411 & 127.2 & D1, D2 \\
Expert-only-EDL    & 256  & No  & No  & Yes & No  & EDL & 2.121 & 112.7 & D2 \\
Full-CE            & 256  & Yes & Yes & Yes & Yes & CE  & 2.438 & 129.9 & D2 \\
Full-EDL           & 256  & Yes & Yes & Yes & Yes & EDL & 2.438 & 129.9 & D1, D2 \\
\bottomrule
\end{tabular}}
\parbox{\linewidth}{\scriptsize\vspace{2pt}Width is the residual-bottleneck
width for Slim/Simple and the per-expert hidden width for Expert-only/Full.
Parameter counts include the classifier. MACs count convolutional and linear
operations for a $224\times224$ input and exclude normalisation and activation.
Expert-only physically deletes NALA, the auxiliary probe, shared fallback, and
the reliability gate. Expert-only-CE was not evaluated; the matrix is therefore
not a complete structure-by-objective factorial.}
\end{table}

Dataset~2 served as the primary controlled audit and included six candidates
evaluated under ten matched seeds: Slim-CE, Simple-CE, Simple-EDL,
Expert-only-EDL, Full-CE, and Full-EDL. Dataset~1 provided the second-dataset
internal replication and included Simple-CE, Simple-EDL, and Full-EDL under the
same ten-seed protocol. Slim-CE was a prespecified secondary lower-capacity
operating point, whereas Simple-CE and Simple-EDL used the capacity-matched
width-1280 structure. Because Expert-only-CE and several Dataset~1 combinations
were not evaluated, the candidate matrix was not a complete
structural-by-objective factorial. Accordingly, all contrasts and
interpretations were restricted to the combinations that were directly
evaluated.

\subsection{Simple and Slim delivery candidates}\label{sec:simple}

Let $\mathbf{h}\in\mathbb{R}^{576\times H\times W}$ denote the high-level
feature map produced by the MobileNetV3-Small backbone. For hidden width $w$,
the simplified residual bottleneck was
\begin{equation}
B_w(\mathbf{h})
=\operatorname{GN}_2\!\left(
C_{1\times1}^{w\rightarrow576}
\left[
\operatorname{ReLU}\!\left(
\operatorname{GN}_1\!\left(
C_{1\times1}^{576\rightarrow w}[\mathbf{h}]
\right)\right)\right]\right),
\label{eq:simple-bottleneck}
\end{equation}
and its output was
\begin{equation}
\widetilde{\mathbf{h}}
=\mathbf{h}+s\,B_w(\mathbf{h}),
\qquad s=\operatorname{softplus}(a)>0.
\label{eq:simple-output}
\end{equation}
The resulting feature was passed through global average pooling and a ten-class
linear classifier to produce logits. Simple used $w=1280$, whereas the
secondary Slim-CE candidate used $w=256$. Simple-CE and Simple-EDL shared this
architecture and differed only in their training objective. Slim was evaluated
only with CE. These candidates contained no NALA, auxiliary probe, router,
experts, shared fallback, or reliability gate.

The width-1280 Simple design was chosen to approximate the parameter and
computation scale of Full. This limited gross capacity reduction as an
alternative explanation for structural differences. At an input resolution of
$224\times224$, Simple contained approximately 2.411 million parameters and
required 127.2 million multiply--accumulate operations (MACs). Full contained
2.438 million parameters and required 129.9 million MACs, so the respective
reductions were approximately 1.1\% and 2.1\%. The comparison therefore concerns
structural simplification, not substantial model compression. Simple
removes routing, multiple expert paths, shared fallback, NALA, the auxiliary
probe, and the batch-dependent reliability gate. Slim-CE
used $w=256$, matching the hidden width of one original expert, and served only
as the secondary lower-capacity operating point. The residual
bottleneck itself was conventional and was not treated as a standalone
architectural contribution.

\subsection{Expert-only and Full structural candidates}\label{sec:full}

Expert-only-EDL used the same ImageNet-pretrained MobileNetV3-Small backbone. It
retained the evidential router and four expert blocks used by Full, while
physically removing NALA, the auxiliary evidential probe, shared fallback, and
the F1--F4 reliability gate. Given the high-level feature map $\mathbf{h}$, the
router operated on its global-average-pooled representation and produced sparse
top-2 routing weights $\widetilde{w}_i$. Each expert $E_i$ used a
$576\rightarrow256\rightarrow576$ bottleneck constructed from $1\times1$
convolutions, group normalisation, and ReLU. The routed feature was
\begin{equation}
\mathbf{o}_{\mathrm{expert}}
=\sum_{i=1}^{4}\widetilde{w}_iE_i(\mathbf{h}),
\qquad
\widetilde{\mathbf{h}}
=\mathbf{h}+s_{\mathrm{moe}}\mathbf{o}_{\mathrm{expert}},
\quad s_{\mathrm{moe}}>0.
\label{eq:expert-only-output}
\end{equation}
The model retained the router load-balancing term and was trained only with the
EDL objective. Deleted modules were absent from the module graph and left no
stored or trainable placeholder parameters. Because several components were
removed together, this candidate audited the combined structural chain. It
could not isolate the contribution of an individual deleted component.

Full was the motivated maximal structural candidate evaluated in the audit. The
MobileNetV3-Small backbone was divided into low- and high-level stages, with
NALA operating on the intermediate low-level feature before the high-level
stage produced $\mathbf{h}$. An auxiliary evidential probe read $\mathbf{h}$.
The evidential router and four experts produced
$\mathbf{o}_{\mathrm{expert}}$, and a separate shared branch produced
$\mathbf{o}_{\mathrm{shared}}$. A parameter-free sample-level reliability gate
$g\in[0,1]$, broadcast over the feature map, interpolated between these two
paths. The fused feature was
\begin{equation}
\mathbf{h}_{\mathrm{out}}
=\mathbf{h}
+s_{\mathrm{moe}}
\left[
\mathbf{o}_{\mathrm{shared}}
+g\left(
\mathbf{o}_{\mathrm{expert}}
-\mathbf{o}_{\mathrm{shared}}
\right)
\right],
\qquad s_{\mathrm{moe}}>0.
\label{eq:full-fusion}
\end{equation}
Thus, $g=0$ selected the shared output within the adaptive branch and $g=1$
selected the routed expert output, while the outer residual preserved the
backbone feature in both cases. Global average pooling and a ten-class linear
classifier produced the final logits. Full-CE and Full-EDL shared this
structural graph and differed in their specified training objectives. The
complete Full-EDL graph and its two principal internal modules are provided in
Appendix Figs.~\ref{fig:app-full}--\ref{fig:app-ugmoe}.

Within Full, NALA operated on the intermediate low-level feature
$\boldsymbol{\ell}$. Two $1\times1$ projections produced $\mathbf{Q}$ and
$\mathbf{K}$, from which the attention branch computed
\begin{equation}
\mathbf{A}_{\mathrm{att}}
=\sigma\!\left[
W_o\!\left(
\operatorname{ReLU}(\mathbf{Q})\odot\operatorname{ReLU}(\mathbf{K})
+\operatorname{ELU}(\mathbf{Q})\odot\operatorname{ELU}(\mathbf{K})
\right)\right].
\label{eq:nala-attention}
\end{equation}
A separate $3\times3$ convolution--BatchNorm--ReLU--$1\times1$ convolution
pathway produced a label-free suppression map $\mathbf{N}\in[0,1]$ through a
sigmoid. The spatial reliability map and residual output were
\begin{equation}
\mathbf{G}
=\mathbf{A}_{\mathrm{att}}\odot(1-\mathbf{N}),
\qquad
\boldsymbol{\ell}'
=\boldsymbol{\ell}
+s_{\mathrm{NALA}}
\left(
\boldsymbol{\ell}\odot\mathbf{G}_{\mathrm{broadcast}}
\right),
\quad s_{\mathrm{NALA}}>0.
\label{eq:nala-reliability}
\end{equation}
The suppression map received no pixel-level supervision and modulated only the
residual feature contribution, while the identity path remained intact.
Accordingly, $\mathbf{N}$ was treated as an internal spatial-weighting signal
rather than a verified estimate of physical noise or artefacts.

The parameter-free reliability gate combined four sample-level factors. Let
$c=\max_i w_i$ denote the largest normalised router weight before sparse top-2
selection. The two routing-confidence factors were
\begin{equation}
F_1=\sigma\!\left(\frac{c-\tau}{t_g}\right),
\qquad
F_2=\exp[-\beta_c(1-c)],
\label{eq:routing-confidence}
\end{equation}
where $\tau$ and $t_g$ were scheduled threshold and temperature parameters. The
auxiliary probe converted its ten-class logits to Dirichlet strength $S_p$ and
uncertainty $u_p=10/S_p$. Its batch-relative excess uncertainty
$u_p^{\mathrm{rel}}$ produced $F_3=\exp(-\beta_pu_p^{\mathrm{rel}})$. Similarly,
router strength $S_r$ defined $u_r=4/S_r$, and normalised excess router
uncertainty $u_r^{\mathrm{rel}}$ produced
$F_4=\exp(-\beta_ru_r^{\mathrm{rel}})$. The pre-floor gate was
\begin{equation}
g_0=F_1F_2F_3F_4.
\label{eq:prefloor-gate}
\end{equation}
Thus, $F_1$ and $F_2$ originated from the same routing-confidence channel, while
$F_3$ and $F_4$ added probe- and router-uncertainty modulation. Each factor lay
in $(0,1]$ after activation, but these signals were not assumed to be
statistically independent. ``Parameter-free'' refers to the absence of a
separate trainable gating network. The upstream router and probe remained
trainable components.

The base interpolation gate was fixed at 1 during the first five training
epochs. After this warm-up, $\tau$ was annealed from 0.30 to 0.45 and $t_g$ from
0.30 to 0.20 over eight epochs. An early floor on $F_1$ decayed from 0.10 to 0
over six epochs. The uncertainty factors $F_2$--$F_4$ became active at
epoch 7 with three-epoch coefficient annealing. For $F_3$, the reference was the
within-batch 70th percentile of probe uncertainty, capped at 0.95. The resulting
factor depended on batch composition and yielded $F_3=1$ for batch-size-one
inference. $F_4$ acted on router uncertainty above the fixed reference
$u_r=0.60$. Probe and router uncertainties used for $F_2$--$F_4$ modulation were
stop-gradient signals where specified in the implementation. After factor
multiplication, the final forward gate was
\begin{equation}
g=\max(g_0,\eta_{\mathrm{post}}),
\label{eq:forward-gate}
\end{equation}
where $\eta_{\mathrm{post}}$ was annealed from 0.08 to 0.03 over ten epochs after
warm-up. A separate training-only gradient floor was annealed from 0.05 to 0.02
over the same period through a straight-through construction. It modified the
backward path but not the forward gate value. The complete coefficient schedule
is provided in Supplementary Table~S1.

\subsection{Learning objectives}\label{sec:objectives}

Slim-CE, Simple-CE, and Full-CE used class-weighted cross-entropy as the primary
classification objective. For a minibatch of size $B$, with
$\mathbf{q}_n=\operatorname{softmax}(\mathbf{z}_n)$ denoting the raw class
probabilities and $\omega_{y_n}$ the effective-number class weight, the primary
loss was
\begin{equation}
\mathcal{L}_{\mathrm{CE}}
=-\frac{1}{B}\sum_{n=1}^{B}
\omega_{y_n}\log q_{n,y_n},
\label{eq:ce-loss}
\end{equation}
where the class weights used $\beta=0.999$. These loss weights were distinct
from the inverse-frequency sample weights used by the training sampler.
Temperature scaling was not part of the training objective and was fitted only
after checkpoint selection. Simple-CE and Slim-CE used the primary loss alone,
whereas the architecture-specific auxiliary terms retained by Full-CE are
defined below.

Simple-EDL, Expert-only-EDL, and Full-EDL used a class-weighted evidential
objective. For $K=10$ classes, evidence, Dirichlet concentration, and total
strength were defined as
\begin{equation}
e_{n,k}=\operatorname{softplus}(z_{n,k}),
\qquad
\alpha_{n,k}=e_{n,k}+1,
\qquad
S_n=\sum_{k=1}^{K}\alpha_{n,k}.
\label{eq:edl-quantities}
\end{equation}
The corresponding internal Dirichlet expectation was $\alpha_{n,k}/S_n$, and
evidential uncertainty was $u_n=K/S_n$. With digamma function $\psi(\cdot)$, the
weighted data term was
\begin{equation}
\mathcal{L}_{\mathrm{ENLL}}
=\frac{1}{B}\sum_{n=1}^{B}
\omega_{y_n}
\left[
\psi(S_n)-\psi(\alpha_{n,y_n})
\right].
\label{eq:enll-loss}
\end{equation}
The primary EDL loss was
\begin{equation}
\mathcal{L}_{\mathrm{EDL}}
=\mathcal{L}_{\mathrm{ENLL}}
+\lambda_t\frac{1}{B}\sum_{n=1}^{B}
D_{\mathrm{KL}}\!\left[
\operatorname{Dir}(\boldsymbol{\alpha}_n)
\,\Vert\,
\operatorname{Dir}(\mathbf{1})
\right],
\qquad
\lambda_t=\min(1,t/10).
\label{eq:edl-loss}
\end{equation}
Effective-number weights used $\beta=0.999$ and applied only to the evidential
data term. The KL coefficient reached 1.0 after ten epochs. Temperature scaling
was not part of EDL training. The Dirichlet expectation and $u_n$ were internal
EDL quantities, whereas cross-objective evaluation used the common
softmax-based conventions defined in Sections~\ref{sec:problem}
and~\ref{sec:metrics}.

The total objective depended on the structural candidate. Slim-CE and Simple-CE
used $\mathcal{L}_{\mathrm{CE}}$ alone, and Simple-EDL used
$\mathcal{L}_{\mathrm{EDL}}$ alone. Expert-only-EDL additionally retained the
router load-balancing term. For four experts, let $I_i$ denote mean soft routing
importance and $P_i$ the hard top-2 load, each normalised to sum to one. We used
\begin{equation}
\mathcal{L}_{\mathrm{LB}}
=4\sum_{i=1}^{4}\left(I_i^2+P_i^2\right).
\label{eq:load-balance}
\end{equation}
Its coefficient increased linearly from zero beginning at epoch 6 and reached
$5\times10^{-4}$ at epoch 10. Full additionally retained an auxiliary probe
loss with coefficient 0.05. Accordingly,
\begin{equation}
\mathcal{L}_{\mathrm{Expert\text{-}only\text{-}EDL}}
=\mathcal{L}_{\mathrm{EDL}}
+\rho_t\mathcal{L}_{\mathrm{LB}},
\label{eq:expert-only-objective}
\end{equation}
and, for $o\in\{\mathrm{CE},\mathrm{EDL}\}$,
\begin{equation}
\mathcal{L}_{\mathrm{Full}\text{-}o}
=\mathcal{L}_{o}
+0.05\,\mathcal{L}_{\mathrm{probe}\text{-}o}
+\rho_t\mathcal{L}_{\mathrm{LB}},
\label{eq:full-objective}
\end{equation}
where $\rho_t$ denotes the annealed load-balancing coefficient. The Full-CE
probe used the CE-form loss, whereas the Full-EDL probe used the EDL-form loss.
In both Full variants, the probe logits were transformed to evidential
uncertainty for the gate computation. Expert-only contained no probe loss, and
Simple, Slim, and NALA received no separate auxiliary supervision.

\subsection{Datasets, partitions, and leakage controls}\label{sec:data}

We evaluated the candidate designs on two in-distribution, ten-class,
multi-organ ultrasound collections assembled from publicly available
image-level datasets, as summarised in Table~\ref{tab:datasets}. Dataset~1
contained
16520 images from three organ groups: breast~\citep{AlDhabyani2020BUSI},
kidney~\citep{KaggleKidneyUS}, and five-stage liver
fibrosis~\citep{Joo2023Liver}. Dataset~2 contained 4403 images from five
organ groups: breast~\citep{Huang2023BUSIWHU}, kidney and pelvic
imaging~\citep{Karim2025MSStateUS}, ovary~\citep{KaggleOvaryUS}, and
thyroid~\citep{Pang2025Thyroid}. For OOD evaluation, Dataset~1 was paired
with an anatomically distinct lung-ultrasound probe containing 2149
images~\citep{Born2020POCUS,Madhu2024POCUS}, whereas Dataset~2 was paired
with a fetal-ultrasound probe containing 1669
images~\citep{Anitha2024Fetal}. Dataset~2 served as the primary
controlled-audit dataset. Dataset~1 provided a second-dataset internal
replication to examine whether the main findings recurred under a different
organ and source composition.

\begin{table}[pos=!htb]
\centering
\caption{Composition and evaluation roles of the two image-level ultrasound
collections. Counts in the source column are shown as number of classes and
images. Tr/Val/Cal/Test denotes the frozen 65/10/5/20\% partition.}
\label{tab:datasets}
\footnotesize
\renewcommand{\arraystretch}{1.12}
\setlength{\tabcolsep}{3.5pt}
\begin{tabular}{@{}p{0.75cm}p{6.1cm}p{2.75cm}p{3.65cm}@{}}
\toprule
\textbf{Set} & \textbf{In-distribution sources} &
\textbf{Tr/Val/Cal/Test} & \textbf{Distribution-shift probe} \\
\midrule
D1 & Breast (3; 781)~\citep{AlDhabyani2020BUSI}; kidney
(2; 9416)~\citep{KaggleKidneyUS}; liver fibrosis
(5; 6323)~\citep{Joo2023Liver}. Total: 10 classes, 16520 images. &
10737 / 1651 / 828 / 3304 &
Lung ultrasound, 2149 images~\citep{Born2020POCUS,Madhu2024POCUS}. \\
\addlinespace[2pt]
D2 & Breast (2; 927)~\citep{Huang2023BUSIWHU}; kidney
(2; 540) and pelvic (2; 625)~\citep{Karim2025MSStateUS}; ovary
(2; 1924)~\citep{KaggleOvaryUS}; thyroid
(2; 387)~\citep{Pang2025Thyroid}. Total: 10 classes, 4403 images. &
2864 / 440 / 219 / 880 &
Fetal ultrasound, 1669 images~\citep{Anitha2024Fetal}. \\
\bottomrule
\end{tabular}
\parbox{\linewidth}{\scriptsize\vspace{2pt}All four partitions are mutually
disjoint at the image-file level. D2 is the primary audit and D1 the internal
replication. The restricted boundary analysis applies D2-trained Full-EDL to
648 breast images from D1 without target fine-tuning or temperature refitting.}
\end{table}

\begin{figure}[pos=!htb]
  \centering
  \includegraphics[width=\linewidth]{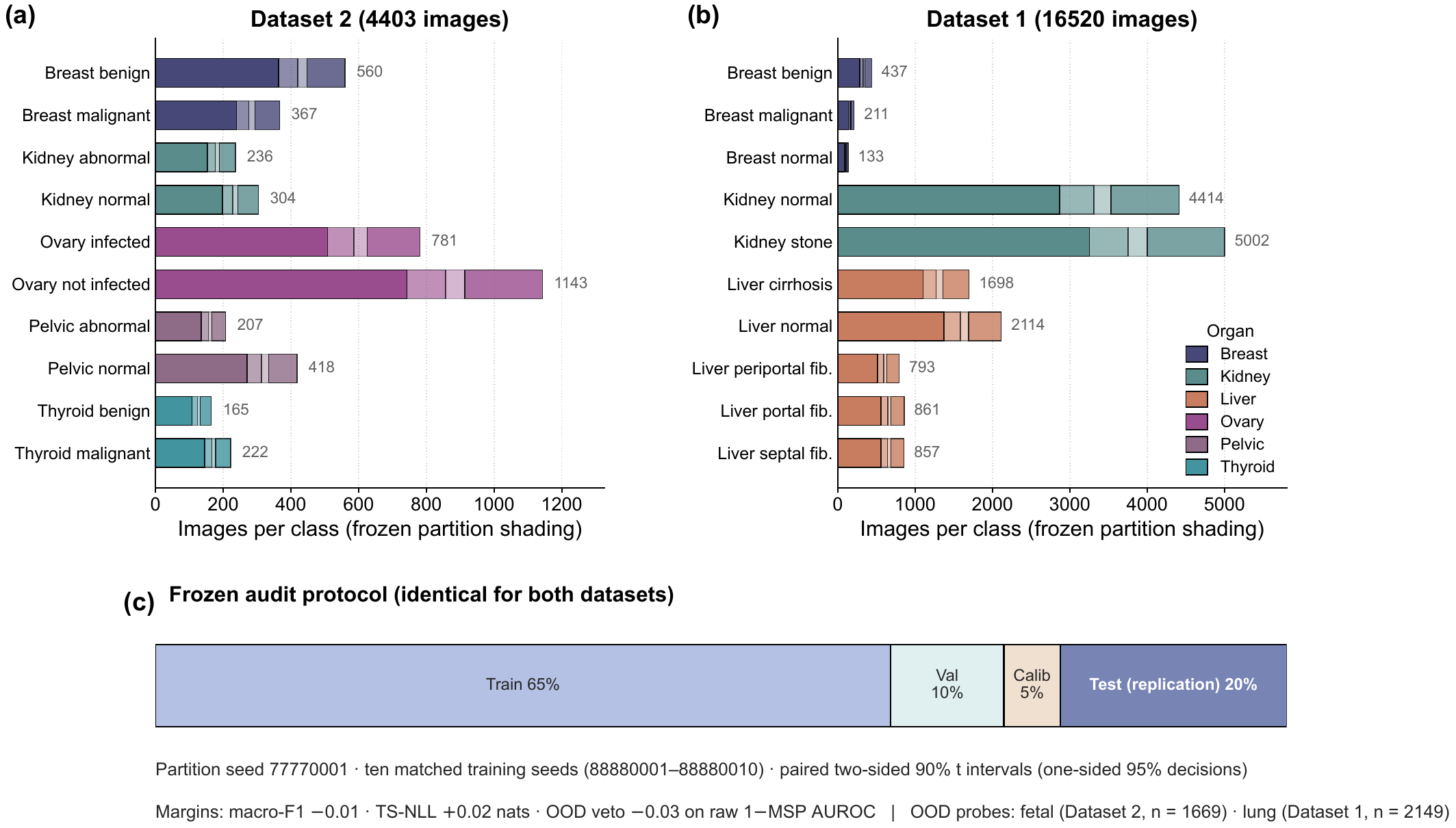}
  \caption{Datasets and frozen audit protocol. Per-class image counts for
  (a) Dataset~2 (4403 images, five organ groups) and (b) Dataset~1
  (16520 images, three organ groups), coloured by organ. The shading within
  each bar shows the frozen 65/10/5/20 image-level partition of that class.
  (c) The shared protocol: fixed partition seed, ten matched training seeds,
  paired two-sided 90\% $t$ intervals (one-sided 95\% decisions), the
  prespecified margins, and the fetal and lung OOD probes, which contribute
  only to the separate OOD decision.}
  \label{fig:workflow}
\end{figure}

Within each in-distribution collection, we created a class-stratified
image-level partition using a fixed partition seed and materialised the
indices before training any candidate model. The partition assigned 65\% of
images to training, 10\% to validation, 5\% to calibration, and 20\% to a
frozen replication test. This produced
10737/1651/828/3304 images for Dataset~1 and
2864/440/219/880 images for Dataset~2, respectively. The four mutually
disjoint partitions served distinct roles. The training subset was used only
for gradient-based parameter optimisation, the validation subset only for
checkpoint selection, and the calibration subset only for fitting a scalar
temperature. The frozen replication test was reserved for final evaluation.
All candidate models used the same partition indices within each dataset. This
enabled matched comparisons. It also ensured that no replication-test image
influenced parameter optimisation, checkpoint selection, or temperature
fitting.
Figure~\ref{fig:workflow} summarises the class composition of both datasets,
the frozen image-level partitions, and the shared decision protocol.

The contributing public sources did not provide verifiable patient, study, or
video identifiers. Partition separation could therefore be enforced only at
the image-file level. The frozen manifests prevented the same image file from
appearing in more than one subset, but unknown dependence among images from the
same patient or study could not be excluded. Source identity was also coupled
to organ or class composition in both in-distribution collections. Holding out
an entire source would therefore remove complete target classes and turn the
evaluation into an unseen-class problem. We limit the primary-audit and
second-dataset replication claims to the frozen image-level partitions.
Patient-aware and same-label source-held-out evaluation remain priorities for
future validation.

As a supporting boundary analysis, we conducted a restricted zero-shot breast
stress test using the ten Dataset~2-trained Full-EDL checkpoints and 648 benign
or malignant breast images from Dataset~1. Before evaluation, we screened the
two breast source collections for cross-dataset duplication and found no
duplicate or near-duplicate images, as documented in Section~S4 of the
Supplementary Material. Each checkpoint was applied to
the target images without target-domain fine-tuning. Its scalar temperature,
fitted on the Dataset~2 calibration partition, was used unchanged and was not
refitted with target labels. The original ten-class output space was retained.
The two breast logits were not extracted and renormalised as a new binary
classifier. We report restricted benign/malignant macro-F1,
restricted two-class accuracy, the cross-organ misclassification rate, raw
ECE, and ECE after applying the unchanged source-domain temperature.

\subsection{Training and optimiser verification}\label{sec:training}

All images were resized to $224\times224$ pixels and normalised using standard
ImageNet channel statistics. Training images additionally underwent random horizontal
flipping with probability 0.5 and random rotation within $\pm10^\circ$.
Validation, calibration, frozen replication-test, external, and OOD images
received only deterministic resizing and normalisation. Class imbalance was
handled with a replacement-based \texttt{WeightedRandomSampler} using
inverse-frequency sample weights. The classification losses separately used
effective-number class weights with $\beta=0.999$. Models were trained with a
batch size of 16 for at most 30 epochs under ten matched seeds, using the same
seed set for every primary candidate and Slim-CE. The exact partition and
training seed values are listed in the Supplementary Material. All ten
prespecified seeds were retained, including low-performing runs, so the
reported variation reflects the complete matched-seed experiment, not post-hoc
run selection.

Optimisation used AdamW with a weight decay of $10^{-4}$. Parameters were grouped by
module, and we verified that every trainable parameter belonged to exactly one
group. The ImageNet-pretrained backbone used
a learning rate of $10^{-4}$, while non-router task-specific modules used
$3\times10^{-4}$. For candidates containing a router, router parameters were
frozen during the first two epochs and then trained at $5\times10^{-5}$. All
active learning rates followed cosine decay to 10\% of their initial values.
Training used automatic mixed precision and gradient-norm clipping at 1.0. The
incumbent checkpoint was replaced only when validation macro-F1 improved by
more than $10^{-4}$. Training stopped after eight consecutive epochs without
such improvement. Validation NLL was not used for checkpoint selection. All
candidates followed the same applicable optimisation rules.

To focus the comparison on architectural structure and training objective,
every candidate used the same ImageNet-pretrained MobileNetV3-Small backbone,
with pretrained backbone weights left intact. Before matched-seed training, the
controlled audit fixed the task-specific initialisation scale at $\gamma=4.0$.
Newly introduced convolutional and linear weights were drawn from a
zero-centred normal distribution with standard deviation 0.02. They were
clipped at two standard deviations and then multiplied by $\gamma$. The
corresponding biases were initialised to zero. Where applicable, residual-scale
initial values and
the router initialisation standard deviation were also scaled by $\gamma$.
Candidates otherwise shared the frozen split manifests, preprocessing, matched
seeds, optimisation and checkpoint-selection rules, and calibration procedure.

\subsection{Calibration, metrics, and OOD scores}\label{sec:metrics}

After checkpoint selection, we fitted one positive scalar temperature $T$
separately for each model and seed using only the corresponding 5\% calibration
partition. The fitting objective was calibration-set NLL computed from
$\operatorname{softmax}(\mathbf{z}/T)$, where $\mathbf{z}$ denotes the model
logits. We first evaluated 300 logarithmically spaced temperatures over
$[0.25,4.0]$. If the minimum occurred at either endpoint, the search was
automatically repeated over $[0.05,10.0]$. Validation and frozen
replication-test observations were not used to fit $T$. The same procedure was
applied symmetrically to CE- and EDL-trained checkpoints, and both raw and
temperature-scaled outputs were retained for evaluation.

Macro-F1 was the primary classification metric, with accuracy and per-class
precision, recall, and F1 reported as supporting measures. For per-class
analysis, Wilson 95\% confidence intervals were computed for precision and
recall. Estimates with either reference support or predicted-positive count
below 30 were flagged as low-support and treated as descriptive. Probabilistic
calibration was assessed using NLL, 15-bin ECE with equal-width confidence bins,
and the multiclass Brier score. Each was computed from both raw and
temperature-scaled probabilities. We did not perform class-wise calibration
analysis. Selective prediction was evaluated using the area under the
risk--coverage curve (AURC), with maximum softmax probability as the confidence
score. Raw and temperature-scaled AURC were reported separately because scaling
need not preserve confidence ordering. Lower NLL, ECE, Brier score, and AURC
indicate better performance.

For OOD evaluation, each frozen in-distribution replication set was compared
with its corresponding anatomically distinct OOD probe. The primary score was
raw $1-\mathrm{MSP}$, where $\mathrm{MSP}=\max_k p_k$ was computed from softmax
probabilities obtained from the unscaled logits. OOD observations were treated
as the positive class. Higher scores therefore denoted more OOD-like inputs,
and higher AUROC indicated better separation. To assess dependence on score
definition and post-hoc scaling, we also reported raw predictive entropy,
temperature-scaled $1-\mathrm{MSP}$, and temperature-scaled predictive entropy.
Entropy was defined as $-\sum_k p_k\log p_k$. All candidates used the same softmax-based
score definitions, regardless of whether they were trained with CE or EDL. OOD
observations were not used for training, checkpoint selection, or temperature
fitting.

\subsection{Statistical analysis and decision rules}\label{sec:statistics}

Formal audit decisions were based on candidate-minus-Full-EDL differences
paired across the ten matched training seeds and analysed separately within
each dataset. For each prespecified endpoint, we estimated the mean paired
difference and a two-sided 90\% paired $t$ interval. The relevant interval bound
corresponds to the prespecified one-sided 95\% decision bound. These rules were
fixed before the internal-replication outcomes were examined. Formal labels were
assigned only to prespecified comparisons. Other comparisons were supporting or
descriptive. The intervals characterise variation across matched training seeds
under the frozen image-level partitions. They are not patient- or source-level
population intervals.

For task performance, we defined the paired difference as
\begin{equation}
\Delta_{\mathrm{F1}}
=\mathrm{F1}_{\mathrm{candidate}}
-\mathrm{F1}_{\mathrm{Full\text{-}EDL}},
\label{eq:delta-f1}
\end{equation}
with a prespecified non-inferiority margin of $-0.01$ macro-F1. This margin
represented the maximum tolerated reduction relative to Full-EDL. Within each
dataset, a comparison was labelled \texttt{PASS} when the lower interval bound
exceeded $-0.01$, \texttt{INFERIOR} when the upper bound was below $-0.01$, and
\texttt{INCONCLUSIVE} otherwise. A \texttt{PASS} indicates that the candidate
met this criterion under the matched-seed design. It does not establish
equality, equivalence, or superiority.

For calibrated probabilistic performance, we defined
\begin{equation}
\Delta_{\mathrm{NLL}}
=\mathrm{NLL}_{\mathrm{candidate}}^{\mathrm{TS}}
-\mathrm{NLL}_{\mathrm{Full\text{-}EDL}}^{\mathrm{TS}},
\label{eq:delta-nll}
\end{equation}
with a prespecified margin of $+0.02$. Because lower NLL is better, this margin
represented the maximum tolerated increase in TS-NLL relative to Full-EDL.
Within each dataset, a comparison was labelled \texttt{PASS} when the upper
interval bound was below $+0.02$, \texttt{FAIL} when the lower bound exceeded
$+0.02$, and \texttt{INCONCLUSIVE} otherwise. A \texttt{PASS} indicates that the
candidate met the TS-NLL criterion. It does not by itself establish superior
calibration. Raw and temperature-scaled ECE and Brier score
were treated as supporting calibration endpoints and did not determine this
decision status.

For the OOD safety veto, we defined
\begin{equation}
\Delta_{\mathrm{OOD}}
=\mathrm{AUROC}_{\mathrm{candidate}}^{\mathrm{raw}\;(1-\mathrm{MSP})}
-\mathrm{AUROC}_{\mathrm{Full\text{-}EDL}}^{\mathrm{raw}\;(1-\mathrm{MSP})}.
\label{eq:delta-ood}
\end{equation}
Within each dataset, a comparison was labelled \texttt{VETO} when the lower
paired-interval bound was less than or equal to $-0.03$ and
\texttt{NOT\_VETOED} otherwise. A \texttt{VETO} indicated that the interval
admitted a reduction of at least 0.03 in the primary OOD AUROC relative to
Full-EDL. A \texttt{NOT\_VETOED} status means only that the veto rule was not
triggered. It does not establish OOD safety, equivalence, or non-inferiority.
Supporting OOD scores were reported for sensitivity but did not
determine or override the primary veto status, which remained separate from the
macro-F1 and TS-NLL decisions.

We also applied an image-level stratified bootstrap within each frozen
replication partition, resampling images within class. This sensitivity analysis
addressed the finite composition of the image-level test set. It neither
replaced the matched-seed analysis nor determined the formal labels. Because
patient, study, and video identifiers were unavailable, image-level resampling
could not capture unknown within-patient or within-study dependence. The
bootstrap is therefore not patient-cluster-aware. Margin-sensitivity analyses
likewise left the primary margins unchanged.

Separately from the ten-seed candidate comparisons, we conducted a supporting
zero-retraining functional audit using three Dataset~2 Full-EDL checkpoints at
$\gamma=4$. Each checkpoint was re-evaluated on the 880-image frozen clean test
set and the 1669-image fetal OOD probe without updating model parameters. To
assess the batch-relative $F_3$ factor, we compared batch sizes of 1, 4, 16,
and 32 under both batch-quantile and fixed-reference modes, recording prediction
flips and probability-level changes. We also evaluated the observed
interpolation gate against two branch counterfactuals, forcing $g=0$ for
shared-only inference and $g=1$ for expert-only inference. Macro-F1, ECE, NLL,
Brier score, AURC, and OOD AUROC were recorded as functional diagnostics. This
three-checkpoint analysis did not determine the formal ten-seed labels. Physical
module deletion was evaluated separately through the retrained Expert-only-EDL
candidate.

\section{Experiments and results}\label{sec:results}

\subsection{Experimental setup and descriptive context}
\label{sec:setup}
Dataset~2 served as the primary controlled audit, with six candidates evaluated
under ten matched seeds: Slim-CE, Simple-CE, Simple-EDL, Expert-only-EDL,
Full-CE, and Full-EDL. Dataset~1 provided a second-dataset internal replication
with Simple-CE, Simple-EDL, and Full-EDL evaluated under the same ten-seed
protocol. The two experiments therefore produced 60 and 30 trained checkpoints,
respectively. Candidates within each dataset shared the frozen image-level
partitions, preprocessing, optimisation,
checkpoint-selection, and temperature-scaling procedures described in
Section~\ref{sec:method}. Candidate comparisons were paired by training seed
and analysed separately within each dataset.

\begin{figure}[pos=!htb]
  \centering
  \includegraphics[width=\linewidth]{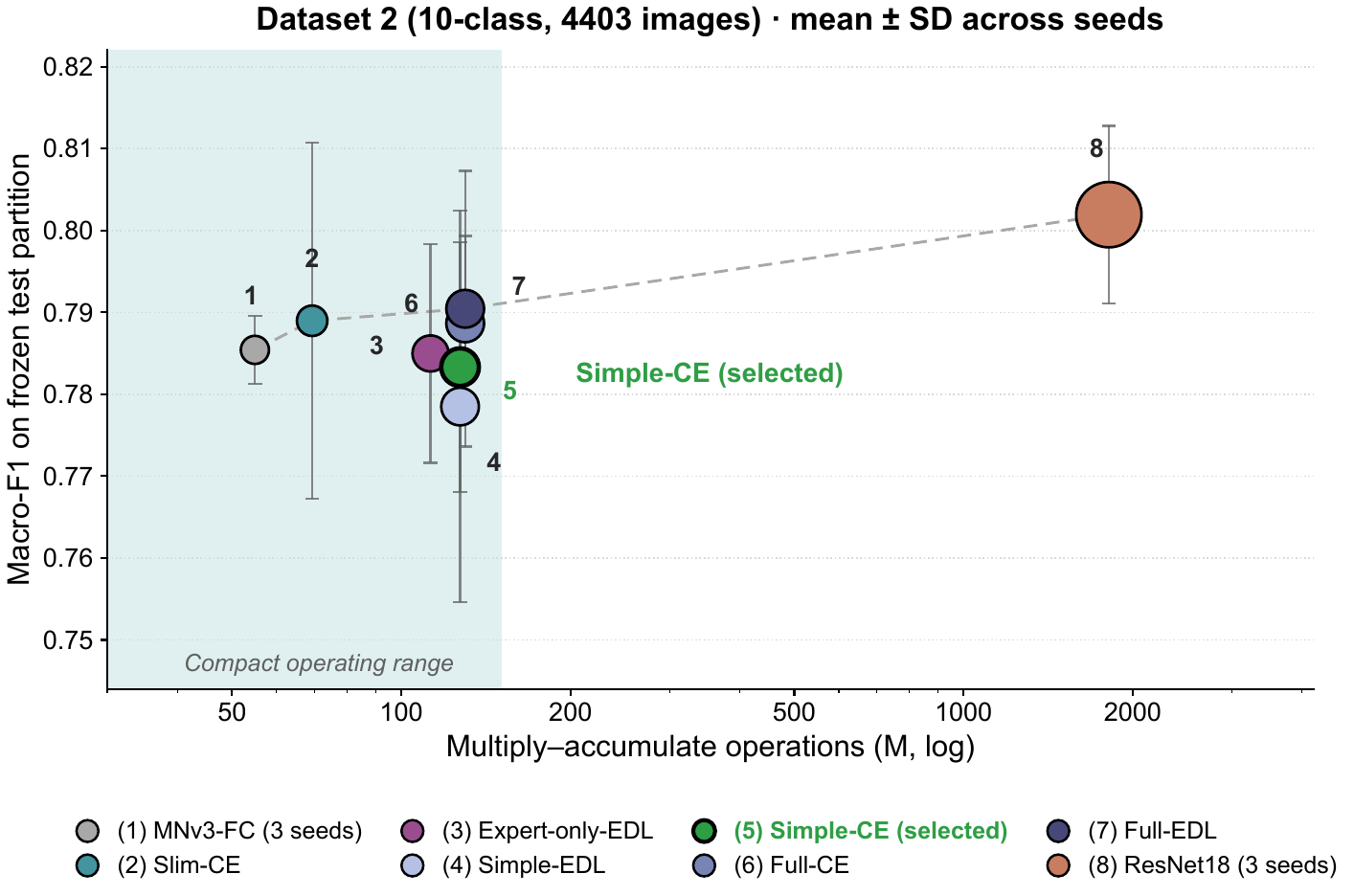}
  \caption{Computational cost versus task performance on Dataset~2. Raw
  macro-F1 (mean $\pm$ SD across seeds) is plotted against multiply--accumulate
  operations on a logarithmic axis. Marker area is proportional to parameter
  count, and the dashed line connects the nondominated mean operating points.
  The shaded band marks the compact operating range. The two contextual
  baselines used three seeds and did not enter formal decisions. Descriptively,
  ResNet18 required approximately fourteen times the computation of the audited
  candidates for a $+0.012$ mean macro-F1 difference over Full-EDL.}
  \label{fig:efficiency}
\end{figure}

For descriptive context, we additionally evaluated two conventional pretrained
CNN classifiers on Dataset~2 under three matched seeds, as reported in
Table~\ref{tab:context-baselines}. MobileNetV3-Small followed by global
average pooling and a linear classifier contained 0.933 million parameters and
required 54.9 million MACs, yielding a macro-F1 of $0.785\pm0.004$, a TS-ECE
of 0.036, and a TS-NLL of 0.392. The corresponding ResNet18 classifier
contained 11.182 million parameters and required 1813.6 million MACs, yielding
a macro-F1 of $0.802\pm0.011$, a TS-ECE of 0.029, and a TS-NLL of 0.377.
These baselines provided standard-CNN context only and were not included in the
ten-seed paired decisions or OOD analysis. Figure~\ref{fig:efficiency}
positions all candidates and both context baselines by computational cost and
macro-F1.

\begin{table}[pos=!htb]
\centering
\caption{Conventional pretrained-CNN context on Dataset~2. Values are
mean$\pm$sample standard deviation over three matched seeds.}
\label{tab:context-baselines}
\small
\renewcommand{\arraystretch}{1.08}
\setlength{\tabcolsep}{5pt}
\begin{tabular*}{\linewidth}{@{\extracolsep{\fill}}lrrrrrr@{}}
\toprule
\textbf{Classifier} & \textbf{Seeds} & \textbf{Params (M)} &
\textbf{MACs (M)} & \textbf{Macro-F1} & \textbf{TS-ECE} & \textbf{TS-NLL} \\
\midrule
MobileNetV3-Small--FC & 3 & 0.933 & 54.9 & $0.785\pm0.004$ & 0.036 & 0.392 \\
ResNet18--FC           & 3 & 11.182 & 1813.6 & $0.802\pm0.011$ & 0.029 & 0.377 \\
\bottomrule
\end{tabular*}
\parbox{\linewidth}{\scriptsize\vspace{2pt}These models provide descriptive
context only. They were not included in the ten-seed paired decisions or OOD
analysis.}
\end{table}

\subsection{Dataset~2 primary controlled audit}\label{sec:d2-audit}
\begin{figure}[pos=!htb]
  \centering
  \includegraphics[width=\linewidth]{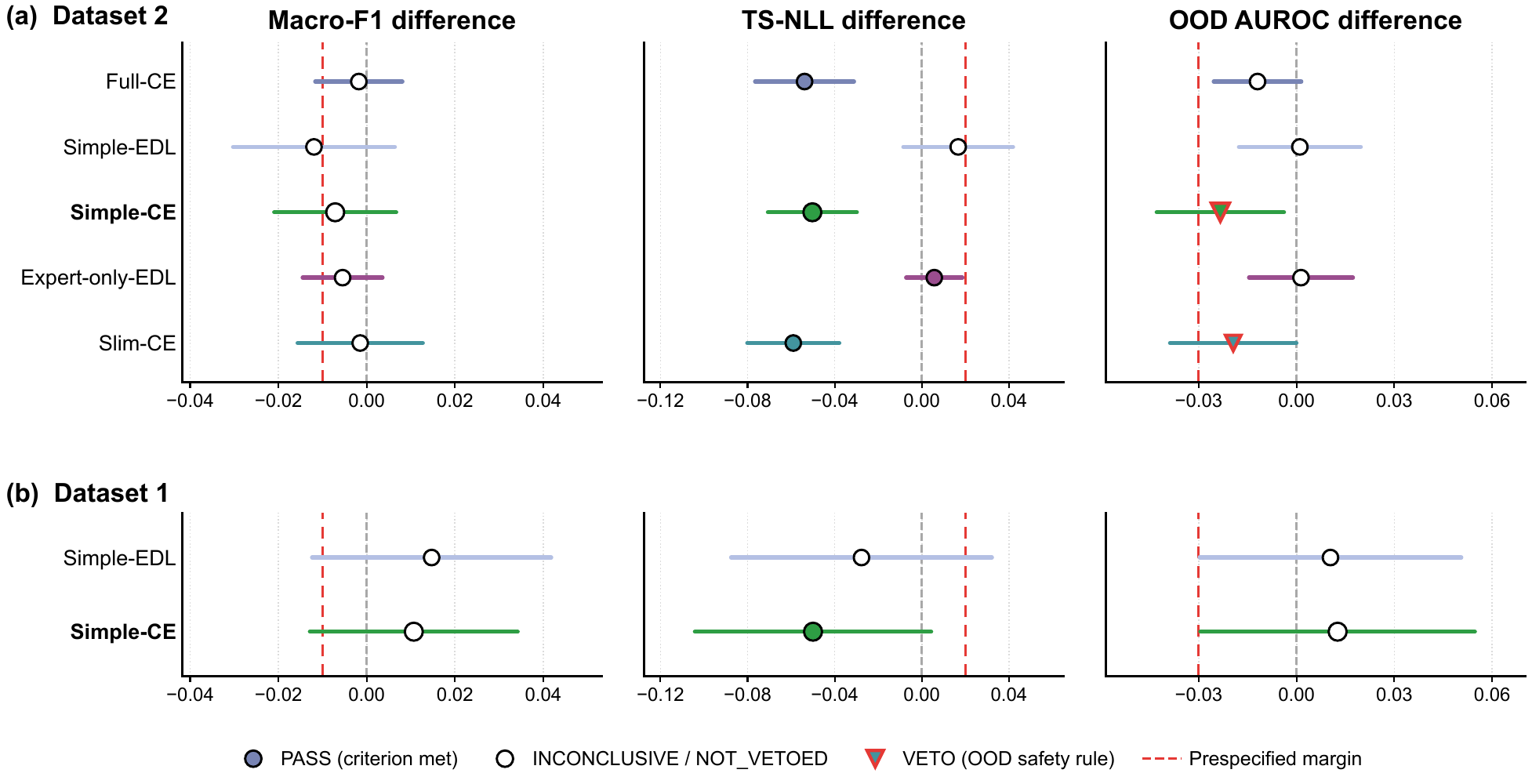}
  \caption{Paired candidate-minus-Full-EDL effects on (left) macro-F1,
  (centre) temperature-scaled negative log-likelihood (TS-NLL), and (right)
  primary raw $1-\mathrm{MSP}$ OOD AUROC for (a) Dataset~2 and (b) Dataset~1.
  Points show paired mean differences and horizontal bars show two-sided 90\%
  paired $t$ intervals across ten matched seeds. Grey dashed lines mark zero.
  Red dashed lines mark the prespecified F1 ($-0.01$), TS-NLL ($+0.02$), and
  OOD-veto ($-0.03$) margins. Filled markers denote \texttt{PASS}, open
  markers denote \texttt{INCONCLUSIVE} or \texttt{NOT\_VETOED}, and red-edged
  downward triangles denote \texttt{VETO}. Interval colour identifies each
  candidate consistently across figures. Decisions are interpreted separately
  for each endpoint and dataset.}
  \label{fig:effects}
\end{figure}

As shown in Table~\ref{tab:d2-results} and Fig.~\ref{fig:effects}, mean
macro-F1 across the six Dataset~2 candidates ranged from 0.778 to 0.790.
Full-EDL had the
nominally highest mean at $0.790\pm0.017$, but the differences from the
alternatives were small relative to matched-seed variation. All five
candidate-minus-Full-EDL macro-F1 comparisons were \texttt{INCONCLUSIVE}.
Their paired mean differences ranged from $-0.0119$ for Simple-EDL to $-0.0015$
for Slim-CE, and no 90\% interval supported either \texttt{PASS} or
\texttt{INFERIOR}. Thus, Dataset~2 showed no established macro-F1 gain from
retaining Full-EDL. The evidence was also insufficient to declare an
alternative non-inferior under the $-0.01$ margin.

The calibrated-loss results separated more clearly by training objective.
Full-EDL had a TS-NLL of $0.427\pm0.024$, whereas Full-CE, Simple-CE, and
Slim-CE achieved $0.373\pm0.027$, $0.376\pm0.019$, and $0.368\pm0.023$,
respectively. All three CE candidates received a \texttt{PASS} decision, with
paired differences of $-0.0538$ [${-0.0765},{-0.0311}$], $-0.0503$
[${-0.0707},{-0.0299}$], and $-0.0590$ [${-0.0801},{-0.0379}$], respectively.
Expert-only-EDL also met the TS-NLL criterion, with a difference of $+0.0057$
[${-0.0072},{+0.0186}$]. Simple-EDL remained \texttt{INCONCLUSIVE} at
$+0.0167$ [${-0.0085},{+0.0419}$]. Supporting TS-ECE means were 0.025--0.031
for the CE candidates and 0.039--0.044 for the EDL candidates, but ECE did not
determine the formal decisions.

\begin{figure}[pos=!htb]
  \centering
  \includegraphics[width=\linewidth]{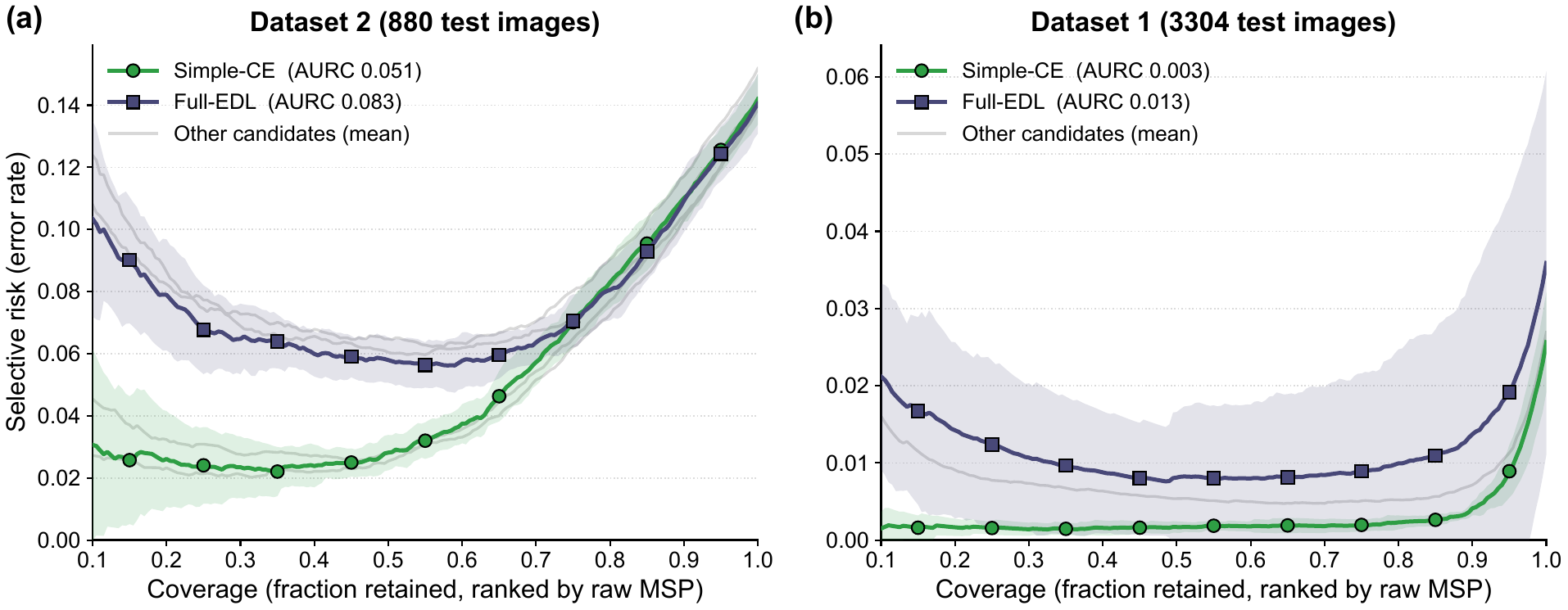}
  \caption{Raw-MSP risk--coverage curves for (a) Dataset~2 and (b) Dataset~1.
  Curves show the mean selective risk across the ten matched seeds as coverage
  varies under confidence-ranked referral. Shaded bands show $\pm$1 SD for
  Simple-CE and Full-EDL, and thin grey lines show the remaining candidates.
  Legend values report mean raw-MSP AURC. Simple-CE maintained lower mean
  selective risk than Full-EDL over most of the coverage range on both
  datasets. These curves are supporting descriptive evidence, not formal
  decision criteria.}
  \label{fig:riskcov}
\end{figure}

Selective-prediction performance showed a similar separation by objective.
Using raw maximum softmax probability as the confidence score, Full-EDL had an
AURC of $0.083\pm0.011$, while Full-CE, Simple-CE, and Slim-CE achieved
$0.052\pm0.011$, $0.051\pm0.007$, and $0.049\pm0.010$, respectively.
Simple-EDL and Expert-only-EDL had AURCs of $0.088\pm0.014$ and
$0.088\pm0.011$, respectively. Because lower AURC indicates better
risk--coverage ordering, the CE-trained candidates had lower mean AURC under the
raw-MSP referral rule. AURC remained a supporting selective-prediction endpoint,
not a probability-calibration measure or formal decision criterion.
Figure~\ref{fig:riskcov} shows the underlying risk--coverage curves for both
datasets.

The primary raw $1-\mathrm{MSP}$ OOD endpoint introduced a different
constraint. Full-EDL achieved an AUROC of $0.960\pm0.027$, compared with
$0.936\pm0.019$ for Simple-CE and $0.940\pm0.015$ for Slim-CE. Their paired
differences were $-0.0234$ [${-0.0429},{-0.0039}$] and $-0.0194$
[${-0.0388},{-0.0001}$], respectively, and both triggered the
\texttt{VETO} because their lower interval bounds were less than or equal to
$-0.03$. Full-CE, Simple-EDL, and Expert-only-EDL received
\texttt{NOT\_VETOED} decisions. Their mean AUROCs were 0.948, 0.961, and 0.961,
respectively. These statuses apply to the Dataset~2 raw $1-\mathrm{MSP}$
endpoint. Under this rule, the two vetoes ruled out Simple-CE and Slim-CE as
OOD-safe replacements for Full-EDL.

\begin{table}[pos=!htb]
\centering
\caption{Dataset~2 primary controlled-audit results. Values are
mean$\pm$sample standard deviation across ten matched seeds. ECE and NLL are
shown as raw/temperature-scaled (TS) values.}
\label{tab:d2-results}
\scriptsize
\renewcommand{\arraystretch}{1.10}
\setlength{\tabcolsep}{2.2pt}
\begin{tabular*}{\linewidth}{@{\extracolsep{\fill}}lccccccc@{}}
\toprule
\textbf{Candidate} & \makecell{\textbf{Macro-F1}\\\textbf{(F1)}} &
\makecell{\textbf{Raw}\\\textbf{ECE}} & \makecell{\textbf{TS}\\\textbf{ECE}} &
\makecell{\textbf{Raw}\\\textbf{NLL}} & \makecell{\textbf{TS-NLL}\\\textbf{(NLL)}} &
\makecell{\textbf{Raw-MSP}\\\textbf{AURC}} &
\makecell{\textbf{Raw $1-\mathrm{MSP}$}\\\textbf{OOD AUROC (OOD)}} \\
\midrule
Full-EDL & $0.790\!\pm\!0.017$/Ref. & $0.051\!\pm\!0.015$ & $0.039\!\pm\!0.008$ &
$0.437\!\pm\!0.022$ & $0.427\!\pm\!0.024$/Ref. & $0.083\!\pm\!0.011$ & $0.960\!\pm\!0.027$/Ref. \\
Full-CE & $0.789\!\pm\!0.011$/I & $0.085\!\pm\!0.025$ & $0.029\!\pm\!0.007$ &
$0.480\!\pm\!0.086$ & $0.373\!\pm\!0.027$/P & $0.052\!\pm\!0.011$ & $0.948\!\pm\!0.016$/NV \\
Simple-EDL & $0.778\!\pm\!0.024$/I & $0.057\!\pm\!0.015$ & $0.044\!\pm\!0.011$ &
$0.457\!\pm\!0.034$ & $0.443\!\pm\!0.034$/I & $0.088\!\pm\!0.014$ & $0.961\!\pm\!0.013$/NV \\
Simple-CE & $0.783\!\pm\!0.015$/I & $0.078\!\pm\!0.016$ & $0.031\!\pm\!0.006$ &
$0.455\!\pm\!0.066$ & $0.376\!\pm\!0.019$/P & $0.051\!\pm\!0.007$ & $0.936\!\pm\!0.019$/V \\
Expert-only-EDL & $0.785\!\pm\!0.013$/I & $0.060\!\pm\!0.023$ & $0.041\!\pm\!0.010$ &
$0.447\!\pm\!0.025$ & $0.432\!\pm\!0.020$/P & $0.088\!\pm\!0.011$ & $0.961\!\pm\!0.015$/NV \\
Slim-CE & $0.789\!\pm\!0.022$/I & $0.077\!\pm\!0.013$ & $0.025\!\pm\!0.005$ &
$0.447\!\pm\!0.070$ & $0.368\!\pm\!0.023$/P & $0.049\!\pm\!0.010$ & $0.940\!\pm\!0.015$/V \\
\bottomrule
\end{tabular*}
\parbox{\linewidth}{\scriptsize\vspace{2pt}F1 status: I =
\texttt{INCONCLUSIVE}; P = \texttt{PASS}. OOD status: V = \texttt{VETO};
NV = \texttt{NOT\_VETOED}. Ref. identifies Full-EDL, the reference candidate.
The parenthesised header identifies the associated decision status. Decision
labels derive from paired 90\% intervals, not from the displayed
absolute means. \texttt{NOT\_VETOED} does not establish OOD safety.}
\end{table}

\subsection{Functional audit of the routing and gate chain}
\label{sec:gate-audit}
The batch-relative $F_3$ audit revealed little observable sensitivity at the
three evaluated checkpoints. Changing the inference batch size from 1 to 4,
16, or 32 produced no predicted-class flips under either the batch-quantile or
fixed-reference implementation. Relative to batch-size-one inference, the
largest observed change in maximum softmax probability was approximately
0.0034 under the batch-quantile comparisons and 0.0033 under the
fixed-reference comparisons. Mean ECE remained approximately 0.0503
throughout, while NLL, Brier score, and raw $1-\mathrm{MSP}$ OOD AUROC varied
only within 0.4760--0.4761, 0.2224--0.2225, and 0.9427--0.9428, respectively.
At these checkpoints, the batch-relative $F_3$ term had negligible observable
influence on predictions and probability-level metrics. Its behaviour beyond
the tested models, datasets, and batch compositions remains open.

The branch counterfactual likewise showed limited functional separation.
Shared-only and expert-only inference disagreed on the predicted class for only
1.0\% of images, and the mean Jensen--Shannon divergence between their
probability distributions was 0.0013. Across the three checkpoints,
shared-only, expert-only, and observed gated inference achieved macro-F1 values
of $0.786\pm0.016$, $0.788\pm0.016$, and $0.787\pm0.013$, respectively.
Expert-only inference also had lower mean ECE, NLL, and Brier score
than the observed gated output (0.047 versus 0.050, 0.467 versus 0.476, and
0.219 versus 0.223) and a higher entropy-based OOD AUROC (0.973 versus 0.963).
The observed interpolation therefore improved none of the recorded means over
forced expert-only inference at these checkpoints. This zero-retraining test
assessed forward behaviour. It did not replace physical deletion and retraining.

Physical deletion and retraining provided a complementary ten-seed test of the
Full-only component chain. Expert-only-EDL, which retained the router, four
experts, and load-balancing term while removing NALA, the auxiliary probe,
shared fallback, and the $F_1$--$F_4$ gate, achieved a macro-F1 of
$0.785\pm0.013$ compared with $0.790\pm0.017$ for Full-EDL. The paired
difference was $-0.0055$ [${-0.0145},{+0.0036}$] and remained
\texttt{INCONCLUSIVE}. Its TS-NLL difference was $+0.0057$
[${-0.0072},{+0.0186}$] and met the \texttt{PASS} criterion. Mean raw ECE rose
from 0.051 to 0.060. The primary raw $1-\mathrm{MSP}$ OOD difference was $+0.0014$
[${-0.0146},{+0.0173}$] and received a \texttt{NOT\_VETOED} decision. Together
with the fixed-checkpoint counterfactuals, these results showed no stable task or
calibrated-loss benefit from the Full-only chain. Because several modules were
removed jointly, the retraining comparison cannot isolate any one component.

\subsection{Objective and calibration audit}\label{sec:calibration-audit}
\begin{figure}[pos=!htb]
  \centering
  \includegraphics[width=\linewidth]{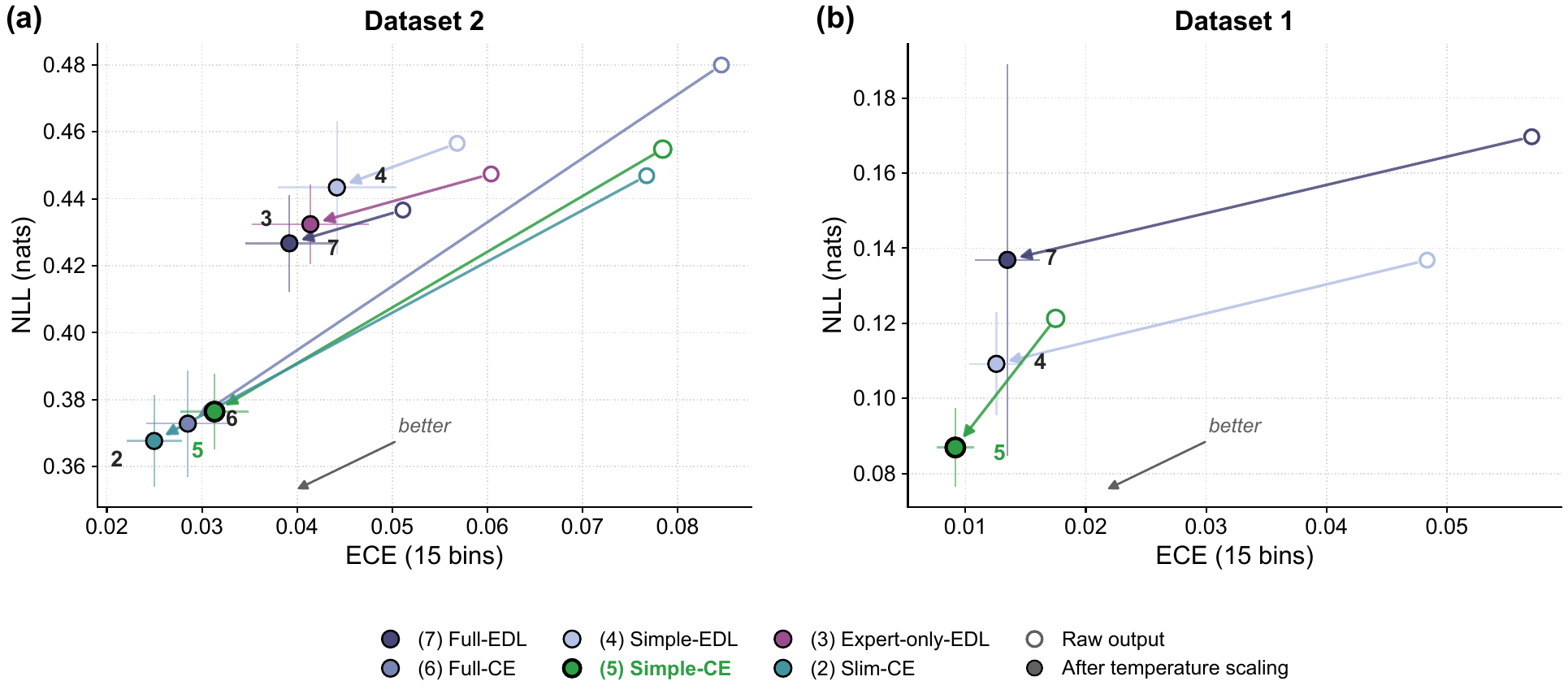}
  \caption{Symmetric post-hoc calibration audit for (a) Dataset~2 and
  (b) Dataset~1. Each candidate is plotted by ECE and NLL. Lower values on both
  axes are preferable. Open markers show raw outputs, filled markers show the
  corresponding temperature-scaled outputs, and arrows connect the two states
  for each candidate. Faint whiskers show two-sided 90\% $t$ intervals for the
  temperature-scaled state across ten seeds. Numbers identify candidates as in
  Fig.~\ref{fig:efficiency}. The apparent raw-calibration advantage of EDL on
  Dataset~2 was not retained after symmetric temperature scaling and did not
  recur on Dataset~1.}
  \label{fig:calibration}
\end{figure}

The raw and temperature-scaled ECE results, shown in
Fig.~\ref{fig:calibration}, illustrated the effect of applying the same
post-hoc calibration opportunity to both objectives. Before temperature
scaling, the three EDL
candidates had raw ECE values of 0.051--0.060, compared with 0.077--0.085 for
the CE candidates. After fitting a separate temperature for each model and seed
on the dedicated calibration partition, the CE candidates achieved TS-ECE
values of 0.025--0.031, whereas the EDL candidates ranged from 0.039 to 0.044.
The apparent raw-ECE advantage of EDL was not retained. The mean ordering
reversed after symmetric temperature scaling. ECE remained a supporting binned
metric. TS-NLL continued to determine the formal calibration decisions.

Dataset~1 provided a contrasting calibration pattern. Full-EDL, Simple-EDL,
and Simple-CE had raw ECE values of $0.057\pm0.023$, $0.048\pm0.008$, and
$0.017\pm0.004$, respectively. After temperature scaling, their ECE values
decreased to $0.013\pm0.005$, $0.013\pm0.004$, and $0.009\pm0.003$. Thus,
unlike on Dataset~2, Simple-CE already had the lowest mean ECE before post-hoc
scaling and retained the lowest mean after scaling. The raw-ECE ordering that
favoured EDL on Dataset~2 therefore did not recur in the Dataset~1 internal
replication, indicating that the apparent objective-level advantage was
dependent on the evaluated dataset and calibration protocol.

\subsection{Dataset~1 internal replication}\label{sec:d1-replication}
On Dataset~1, the nominal macro-F1 ordering differed from that observed on
Dataset~2, as reported in Table~\ref{tab:d1-results} and
Fig.~\ref{fig:effects}. Full-EDL
achieved $0.912\pm0.039$, while Simple-EDL and Simple-CE achieved
$0.927\pm0.014$ and $0.923\pm0.013$, respectively. The paired
candidate-minus-Full-EDL differences were $+0.0147$
[${-0.0124},{+0.0418}$] for Simple-EDL and $+0.0107$
[${-0.0129},{+0.0342}$] for Simple-CE. Both remained
\texttt{INCONCLUSIVE} because their lower bounds fell below the
$-0.01$ margin. The seed standard deviation for Full-EDL was approximately three
times those of the Simple candidates. Thus, Full-EDL showed no established
macro-F1 advantage, and both simplified alternatives remained
\texttt{INCONCLUSIVE}.

The Dataset~1 calibrated-loss results were directionally consistent with the
CE+TS pattern observed on Dataset~2. Full-EDL had a TS-NLL of
$0.137\pm0.090$, compared with $0.109\pm0.023$ for Simple-EDL and
$0.087\pm0.018$ for Simple-CE. Simple-EDL remained \texttt{INCONCLUSIVE},
with a paired difference of $-0.0277$ [${-0.0876},{+0.0322}$]. Simple-CE
received a \texttt{PASS} decision at $-0.0500$ [${-0.1042},{+0.0043}$]. Raw-MSP
AURC was $0.013\pm0.012$, $0.010\pm0.002$, and $0.003\pm0.001$ for Full-EDL,
Simple-EDL, and Simple-CE, respectively. Simple-CE therefore had the lowest
mean AURC in this supporting analysis, consistent with the risk--coverage
curves in Fig.~\ref{fig:riskcov}(b).

The Dataset~1 primary OOD results differed from the Dataset~2 veto pattern.
Under raw $1-\mathrm{MSP}$ scoring against the lung-ultrasound probe, Full-EDL,
Simple-EDL, and Simple-CE achieved AUROCs of $0.920\pm0.068$,
$0.930\pm0.020$, and $0.933\pm0.019$, respectively. The paired differences
were $+0.0104$ [${-0.0297},{+0.0505}$] for Simple-EDL and $+0.0126$
[${-0.0295},{+0.0547}$] for Simple-CE, and both received
\texttt{NOT\_VETOED} decisions. The Simple-CE veto observed with the Dataset~2
fetal probe therefore did not recur with the Dataset~1 lung probe under the same
score and decision rule.

\begin{table}[pos=!htb]
\centering
\caption{Dataset~1 second-dataset internal replication. Values are
mean$\pm$sample standard deviation across ten matched seeds. ECE and NLL are
shown as raw/temperature-scaled (TS) values.}
\label{tab:d1-results}
\scriptsize
\renewcommand{\arraystretch}{1.10}
\setlength{\tabcolsep}{2.2pt}
\begin{tabular*}{\linewidth}{@{\extracolsep{\fill}}lccccccc@{}}
\toprule
\textbf{Candidate} & \makecell{\textbf{Macro-F1}\\\textbf{(F1)}} &
\makecell{\textbf{Raw}\\\textbf{ECE}} & \makecell{\textbf{TS}\\\textbf{ECE}} &
\makecell{\textbf{Raw}\\\textbf{NLL}} & \makecell{\textbf{TS-NLL}\\\textbf{(NLL)}} &
\makecell{\textbf{Raw-MSP}\\\textbf{AURC}} &
\makecell{\textbf{Raw $1-\mathrm{MSP}$}\\\textbf{OOD AUROC (OOD)}} \\
\midrule
Full-EDL & $0.912\!\pm\!0.039$/Ref. & $0.057\!\pm\!0.023$ & $0.013\!\pm\!0.005$ &
$0.170\!\pm\!0.098$ & $0.137\!\pm\!0.090$/Ref. & $0.013\!\pm\!0.012$ & $0.920\!\pm\!0.068$/Ref. \\
Simple-EDL & $0.927\!\pm\!0.014$/I & $0.048\!\pm\!0.008$ & $0.013\!\pm\!0.004$ &
$0.137\!\pm\!0.021$ & $0.109\!\pm\!0.023$/I & $0.010\!\pm\!0.002$ & $0.930\!\pm\!0.020$/NV \\
Simple-CE & $0.923\!\pm\!0.013$/I & $0.017\!\pm\!0.004$ & $0.009\!\pm\!0.003$ &
$0.121\!\pm\!0.030$ & $0.087\!\pm\!0.018$/P & $0.003\!\pm\!0.001$ & $0.933\!\pm\!0.019$/NV \\
\bottomrule
\end{tabular*}
\parbox{\linewidth}{\scriptsize\vspace{2pt}F1 status: I =
\texttt{INCONCLUSIVE}; P = \texttt{PASS}. OOD status: NV =
\texttt{NOT\_VETOED}. Ref. identifies Full-EDL, the reference candidate.
The parenthesised header identifies the associated decision status. Decision
labels derive from paired 90\% intervals, not from the displayed
absolute means. \texttt{NOT\_VETOED} does not establish OOD safety.}
\end{table}

\begin{figure}[pos=!htb]
  \centering
  \includegraphics[width=\linewidth]{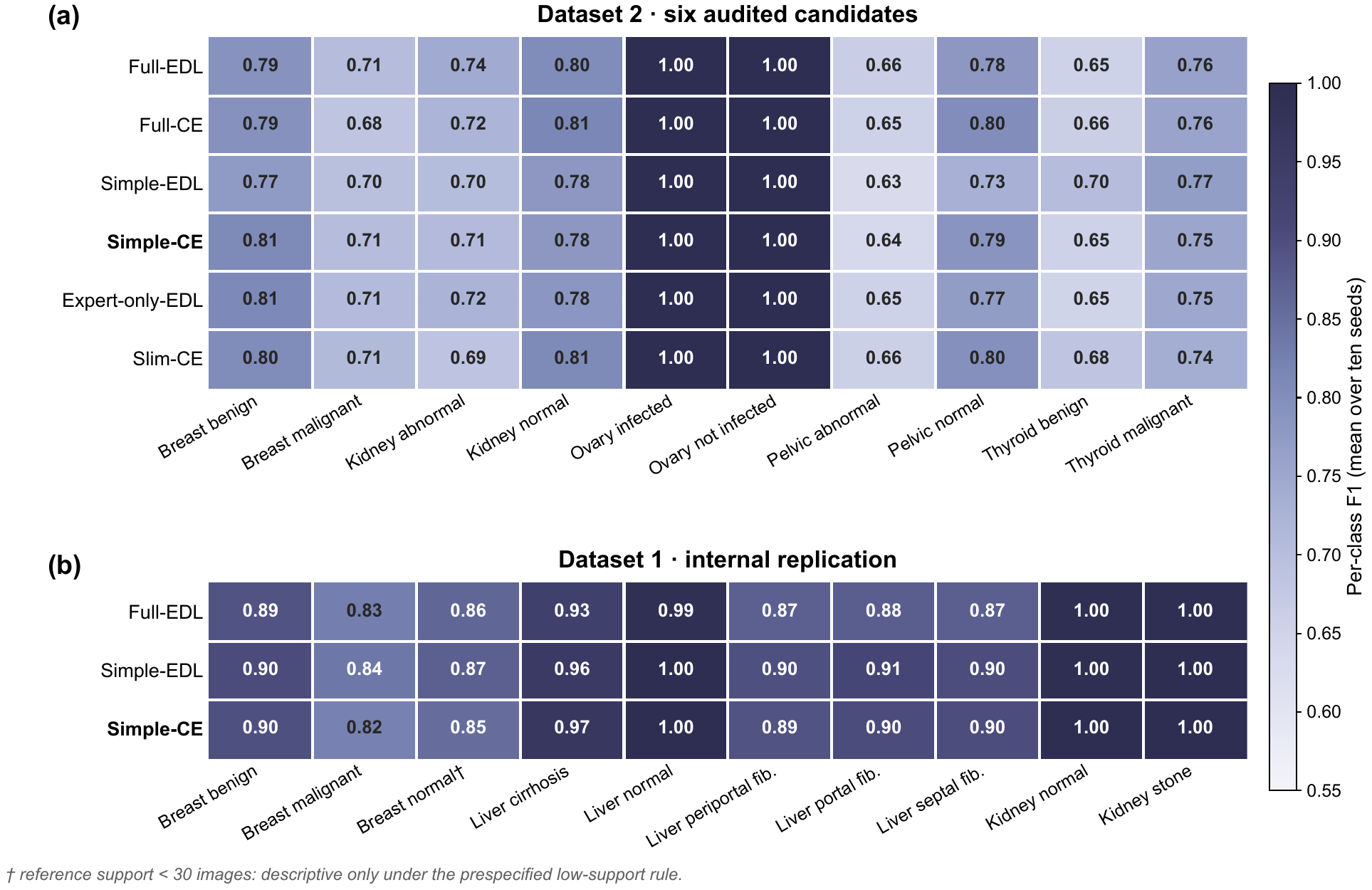}
  \caption{Per-class F1 (mean over the ten matched seeds) for (a) the six
  Dataset~2 candidates and (b) the three Dataset~1 replication candidates.
  Within each dataset, the per-class profiles were similar across candidates.
  The dagger marks a class with reference support below 30 images, reported
  descriptively under the prespecified low-support rule.}
  \label{fig:perclass}
\end{figure}

Figure~\ref{fig:perclass} shows that per-class F1 profiles were similar across
candidates within each dataset and that the same classes remained difficult
for every candidate. Complete seed-level per-class tables are provided in
the Supplementary Material (Section~S6).

\subsection{Cross-dataset synthesis and metric discordance}
\label{sec:synthesis}
As summarised in Fig.~\ref{fig:synthesis}(a), cross-dataset synthesis
separated two recurring findings from two dataset-specific ones. First,
Full-EDL showed no
established macro-F1 gain on either dataset, and the alternatives remained
\texttt{INCONCLUSIVE}. Second, Simple-CE met the TS-NLL criterion and had lower
mean raw-MSP AURC than Full-EDL on both datasets. Two patterns did not recur.
The lower raw ECE of EDL candidates appeared only on Dataset~2, and the
Simple-CE raw $1-\mathrm{MSP}$ OOD veto arose only with the Dataset~2 fetal
probe. The recurring evidence therefore concerned in-distribution task
performance, calibrated loss, and selective-risk ordering. Raw calibration and
OOD separation depended on the dataset and protocol.

The endpoint rankings differed because the metrics assessed distinct behaviours.
Macro-F1 depended on argmax predictions, whereas TS-NLL evaluated the calibrated
probability distribution. Raw-MSP AURC measured the ordering of
in-distribution errors as coverage changed. OOD AUROC instead measured
separation from a specified probe under a specified score.
Simple-CE could therefore meet the TS-NLL criterion and show a lower AURC while
still triggering the Dataset~2 OOD veto. Keeping these endpoints separate
prevented favourable in-distribution calibration or selective-risk results from
being interpreted as evidence of OOD reliability. The resulting model-selection
problem was multi-objective and dependent on the intended operating context. A
single universal ranking would obscure these distinctions.

\subsection{Restricted zero-shot breast stress test}
\label{sec:breast}
\begin{figure}[pos=!htb]
  \centering
  \includegraphics[width=\linewidth]{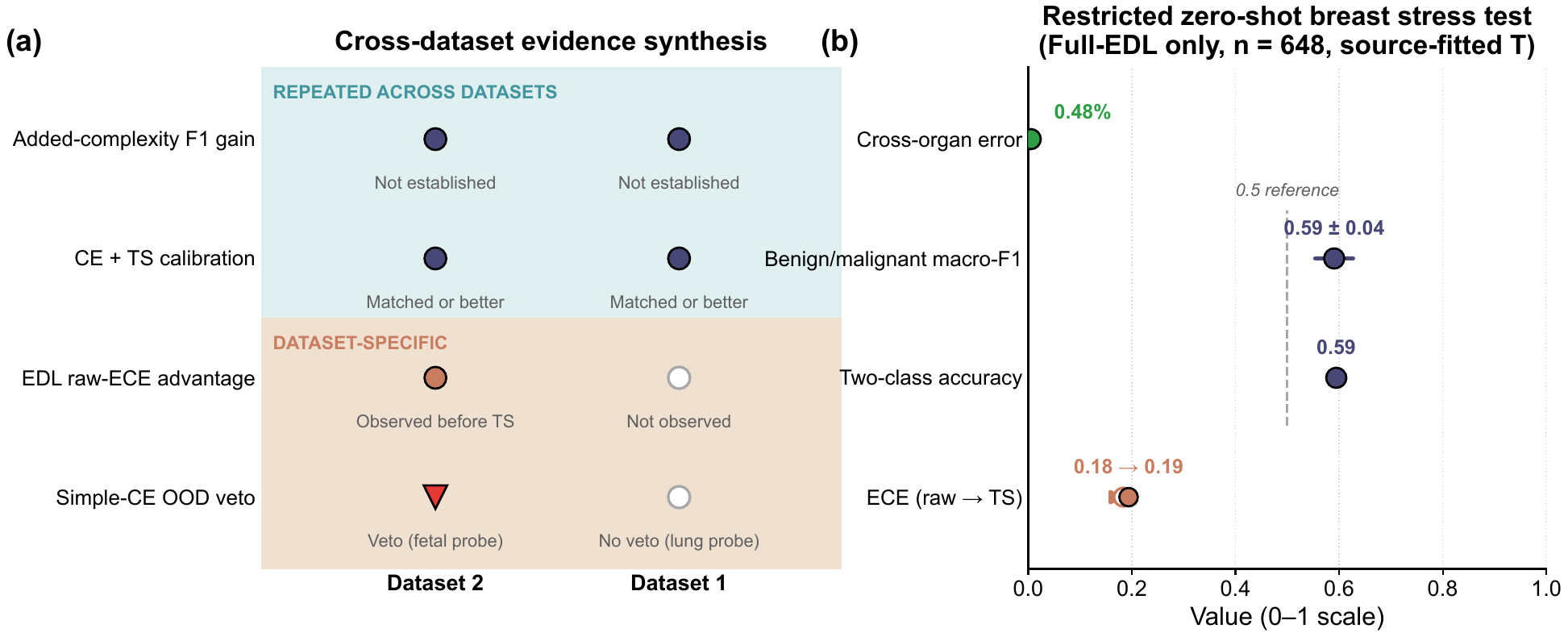}
  \caption{Cross-dataset synthesis and restricted transfer boundary.
  (a) Evidence matrix: the upper band collects findings repeated on both
  datasets (no established macro-F1 gain from added complexity, and CE+TS
  calibration matched or better), and the lower band collects dataset-specific
  findings (the EDL raw-ECE advantage and the Simple-CE OOD veto, marked by
  the red triangle). Open markers denote patterns not observed. (b) Restricted
  zero-shot breast stress test: ten Dataset~2-trained Full-EDL checkpoints
  applied without target fine-tuning or temperature refitting to 648 Dataset~1
  breast images. Coarse organ assignment was largely retained (0.48\%
  cross-organ error), whereas benign/malignant discrimination ($0.59\pm0.04$
  macro-F1 relative to the plotted 0.5 reference) and
  source-domain calibration transfer (ECE 0.18 to 0.19) remained limited.}
  \label{fig:synthesis}
\end{figure}

In the restricted zero-shot breast stress test, shown in
Fig.~\ref{fig:synthesis}(b), the ten Dataset~2-trained Full-EDL checkpoints
were applied to 648 Dataset~1 breast images without target-domain fine-tuning
while retaining the original ten-class output space. The mean cross-organ
misclassification rate was
only 0.48\%, indicating that most predictions remained within the two
breast-related output classes. However, restricted benign/malignant macro-F1
was $0.59\pm0.04$, and restricted two-class accuracy was $0.595\pm0.037$.
This contrast shows that coarse breast-versus-other-organ assignment was largely
retained, whereas fine-grained benign/malignant discrimination remained limited.
The analysis used Full-EDL only and no target adaptation. It is therefore a
boundary test, not external validation of the full candidate set.

Calibration did not transfer with the unchanged source-domain temperature.
Across the ten Full-EDL checkpoints, raw restricted ECE was $0.183\pm0.060$
and changed to $0.194\pm0.058$ after applying the scalar
temperature fitted on the corresponding Dataset~2 calibration partition. No
Dataset~1 breast labels were used to refit the temperature, and no target-domain
calibration was performed. The source-domain temperatures did not improve
calibration on this restricted target subset. Target-fitted calibration was not
evaluated. Together with the classification results, this finding separates
largely retained coarse organ assignment from limited fine-grained diagnosis
and source-domain calibration transfer.

\subsection{Engineering selection}\label{sec:selection}
Considering the endpoints jointly, we selected Simple-CE with post-hoc
temperature scaling for the in-distribution operating objective. Full-EDL
showed no established macro-F1 gain on either dataset. Simple-CE met the TS-NLL
criterion on both datasets and had favourable mean TS-ECE and raw-MSP AURC.
It also replaced the NALA, probe, router, expert, shared-fallback, and
reliability-gate chain with one capacity-matched residual bottleneck. This
simplified the inference graph without relying on substantial parameter
reduction, and its seed variability was lower on Dataset~1. The F1 comparison
nevertheless remained \texttt{INCONCLUSIVE}, and the Dataset~2 OOD veto ruled
out an unconditional replacement. We therefore retain Full-EDL as the audited
maximal candidate and recommend Simple-CE+TS for in-distribution classification,
calibrated probabilities, and selective prediction under this protocol.

\FloatBarrier
\section{Discussion}\label{sec:discussion}

\subsection{Motivated complexity still requires functional evidence}

A central point in this study is that a plausible design rationale does not by
itself establish functional value at the system level. NALA spatial weighting,
the auxiliary evidential probe, sparse expert routing, shared fallback, and
reliability gating each addressed a reasonable source of uncertainty or
heterogeneity in multi-organ ultrasound classification. The system-level
evidence was more restrained. Fixed-checkpoint
counterfactuals revealed little observable contribution from the audited gate,
and physical deletion showed no stable task or calibrated-loss benefit for the
Full-only chain. Capacity-matched comparisons also showed no established
macro-F1 gain from retaining Full-EDL. These audit levels complement a standard
loss ablation: they examine realised forward behaviour, remove modules from the
computational graph, and control gross capacity during retraining. The
engineering lesson is straightforward. Complex routing and uncertainty modules
should be retained when they make an observable contribution under the intended
training and evaluation protocol.

\subsection{Calibration fairness changes model selection}
The calibration audit shows why probabilistic objectives should be compared
both before and after the same post-hoc calibration opportunity. Raw ECE
captured native confidence and favoured EDL on Dataset~2. After a separate
temperature was fitted for every model and seed, however, the mean ordering
reversed \citep{Guo2017Calibration}. The raw-ECE advantage also failed to recur
on Dataset~1. Positive scalar temperature scaling leaves the predicted class
unchanged. It therefore separates task prediction from the degree to which
confidence mismatch can be corrected using held-out data. Raw and scaled results
answer complementary questions: one characterises native output, and the other
supports a symmetric comparison of post-hoc calibrated loss. ECE remains useful
as a binned summary, but TS-NLL, Brier score, and selective risk provide needed
context. Model selection should consider both views instead of relying on raw
ECE alone.

\subsection{Reliability objectives can disagree}
The disagreement among reliability endpoints argues against treating uncertainty
quality as a single model property. Calibration relates predicted probabilities
to outcomes within the evaluated distribution. Selective prediction asks
whether errors are ordered effectively as coverage changes, whereas OOD
detection measures separation from a specified shifted distribution under a
chosen score \citep{Ovadia2019Trust,Geifman2018AURC}. These objectives need not
produce the same ranking. In our audit, Simple-CE met the calibrated-loss
criterion and had lower mean AURC on both datasets. Yet it triggered the primary
OOD veto with the Dataset~2 fetal probe and not with the Dataset~1 lung probe.
The additional score and scaling conventions likewise showed that OOD
conclusions were tied to the evaluation definition rather than inherited
automatically from in-distribution calibration. For intelligent-system
selection, these findings support endpoint-specific acceptance criteria.
Favourable calibration or referral ordering can justify an in-distribution
choice. OOD-sensitive use requires its own validated decision.

\subsection{Recurring and dataset-specific findings}
The second-dataset analysis helped distinguish findings that recurred from
those that remained contingent on the evaluation setting. Despite different
organ and source compositions, both datasets showed no established macro-F1
gain from Full-EDL and supported the calibrated-loss and selective-risk
competitiveness of Simple-CE+TS. By contrast, the raw-ECE advantage of EDL
appeared only on Dataset~2, and the Simple-CE OOD veto differed between the
fetal and lung probes. The restricted breast analysis further delineated this
boundary: Full-EDL outputs largely remained within breast-related classes, but
fine-grained diagnosis and source-domain calibration did not transfer reliably.
Together, these results support an in-distribution engineering choice for the
evaluated multi-organ ultrasound protocol. Raw calibration, OOD separation, and
cross-domain calibration remain dependent on the dataset and assessment design.

\subsection{Limitations and future work}
The principal data boundary is that both multi-organ collections were assembled
from public image-level sources without verifiable patient, study, or video
identifiers. The frozen manifests prevented the same image file from crossing
partitions, but unknown dependence among images from the same patient or study
could not be excluded. Source identity was also coupled to organ and class
composition, so holding out a complete source would remove target classes
and change the task into unseen-class recognition. The paired intervals
characterise repeated optimisation under the frozen image-level partitions.
They do not represent patient-, source-, or centre-level population uncertainty,
and the ten seeds are not independent clinical replications. Low-support
class-level estimates and the image-level bootstrap remain descriptive.
Extending the inference will require patient-aware, same-label multi-source
collections and prospectively defined multicentre evaluation.

The system-level conclusions are limited to the evaluated compact backbone,
ultrasound classification task, and transfer protocol. Gate counterfactuals
provide functional evidence at specific checkpoints, not a causal account of
parameter evolution. The jointly deleted Expert-only candidate also cannot
resolve the individual effects of NALA, the probe, shared fallback, or the
reliability gate. Because Simple and Full had similar parameter and MAC counts,
the practical advantage is a simpler computational graph, not substantial
compression or established acceleration. The OOD analysis covered two
anatomically distinct probes and a limited set of softmax-based score and scaling
conventions. The restricted breast test evaluated Full-EDL only, without target
adaptation or target-domain calibration. Future work should test the audit
framework across broader backbones and tasks, prospectively specified near- and
far-distribution shifts, independently partitioned target-calibration data, and
prospective deployment settings.

\section{Conclusion}\label{sec:conclusion}
Across matched optimisation, capacity-aware comparison, symmetric calibration,
and repeated seeds, Full-EDL showed no established macro-F1 gain. Its audited
reliability gate also had negligible observable influence at the evaluated
checkpoints. We therefore selected Simple-CE with temperature scaling for
in-distribution classification, calibrated probabilities, and selective
prediction. The F1 comparison remained \texttt{INCONCLUSIVE}, and the
Dataset~2 OOD veto ruled out an unconditional replacement, so Full-EDL remains
the maximal reference. More broadly, intelligent-system components should earn
retention through functional and retraining-based evidence. Probabilistic
objectives deserve symmetric calibration opportunities, and reliability under
distribution shift must be validated separately from in-distribution
performance.

\FloatBarrier
\appendix
\renewcommand{\thefigure}{A.\arabic{figure}}
\renewcommand{\theHfigure}{A.\arabic{figure}}
\setcounter{figure}{0}
\section{Architectural details of the maximal candidate}
\label{app:architecture}

The following diagrams document the maximal Full-EDL candidate evaluated in
the architecture audit. They provide implementation details for this reference
design.

\begin{figure}[pos=H,abovefig=0pt,belowfig=0pt,abovecap=2pt,belowcap=0pt]
  \centering
  \includegraphics[width=0.70\linewidth]{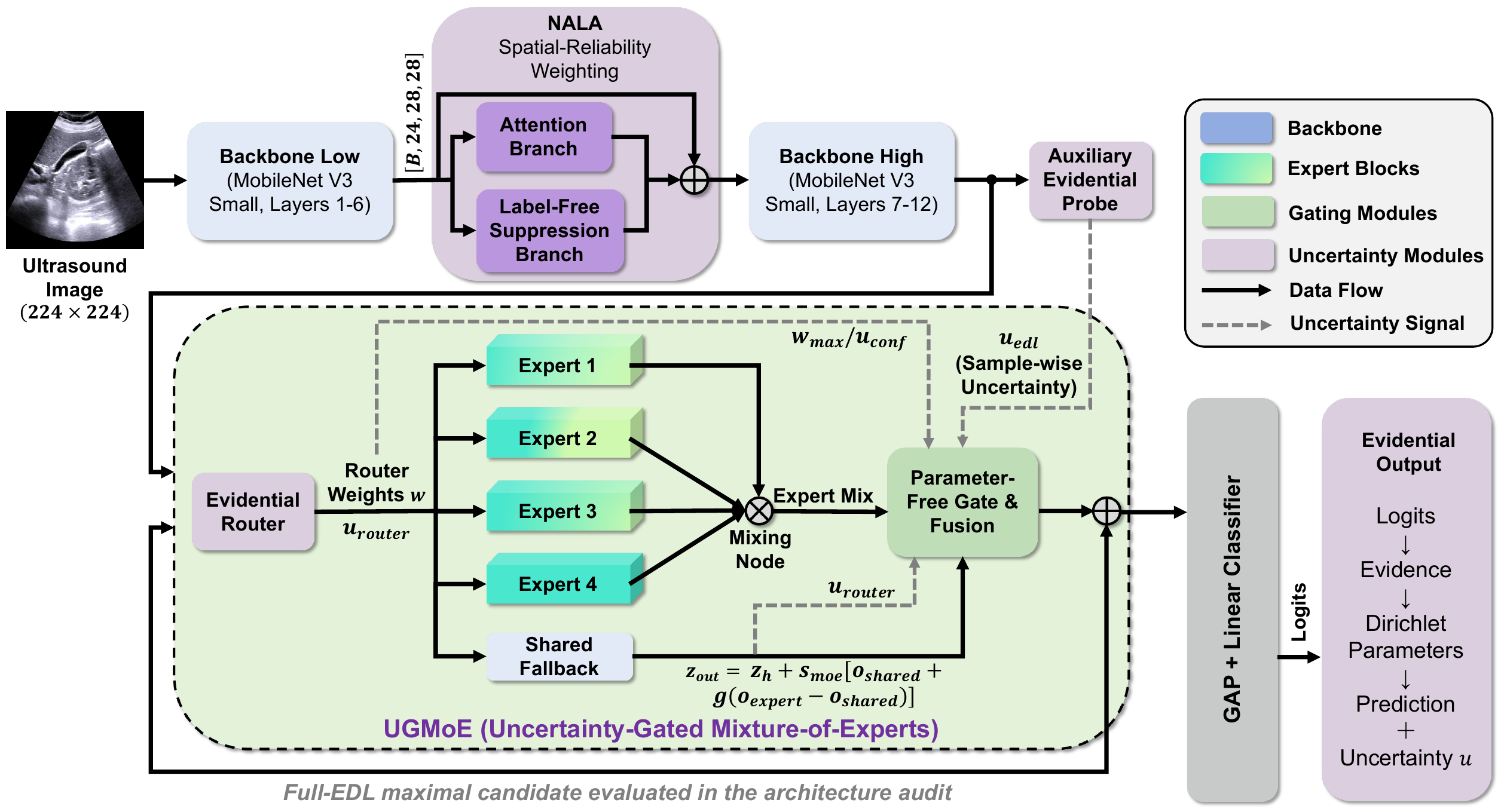}
  \caption{Complete Full-EDL maximal-candidate architecture. MobileNetV3-Small
  is divided into low- and high-level stages. NALA performs intermediate
  spatial-reliability weighting. The auxiliary evidential probe, evidential
  router, four expert blocks, shared fallback, and parameter-free reliability
  gate form the adaptive branch. The gated branch is combined with the
  high-level backbone feature through a positive residual scale before global
  average pooling and evidential output. Full-EDL is shown as the maximal
  reference evaluated in the architecture audit.}
  \label{fig:app-full}
\end{figure}

\begin{figure}[pos=H]
  \centering
  \includegraphics[width=\linewidth]{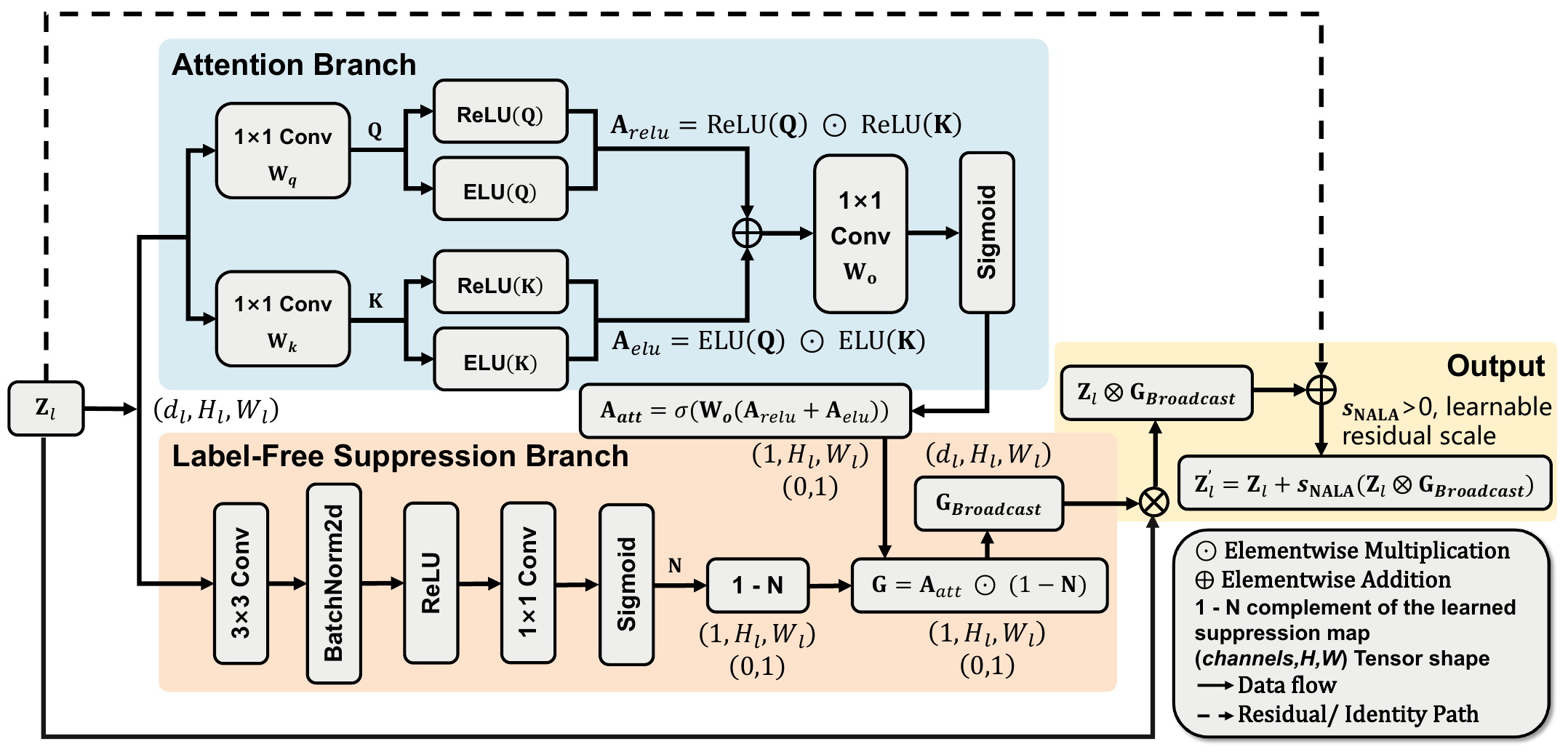}
  \caption{Computational structure of the NALA spatial-reliability module. The
  attention branch combines ReLU- and ELU-based interactions between projected
  query and key features. A separate label-free branch produces a learned
  suppression map. Their product forms a broadcast spatial-reliability map
  that modulates only the positive-scaled residual contribution, leaving the
  identity path intact. The suppression map is an internal weighting signal
  and is not treated as a verified physical noise or artefact estimate.}
  \label{fig:app-nala}
\end{figure}

\begin{figure}[pos=H]
  \centering
  \includegraphics[width=\linewidth]{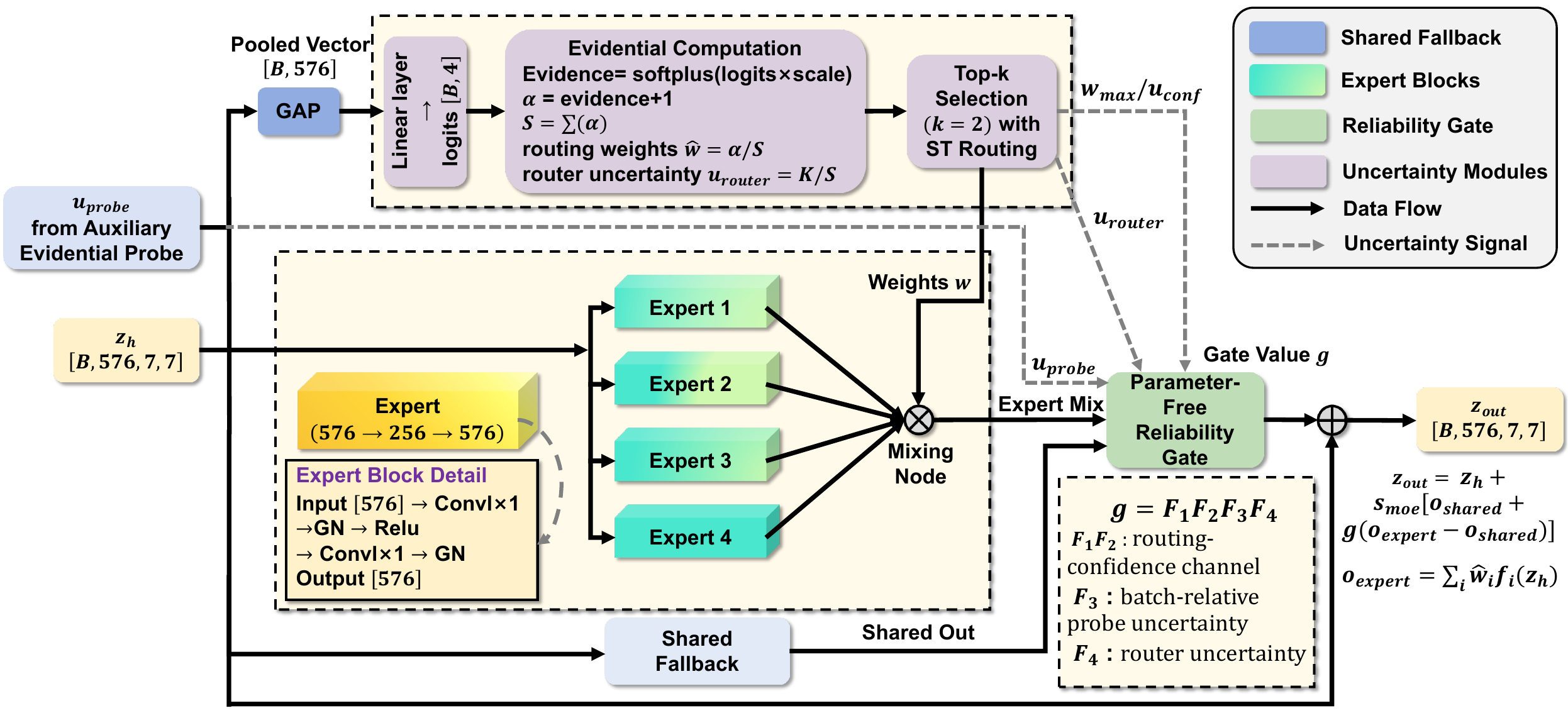}
  \caption{Detailed UGMoE routing, expert mixing, and reliability-gating
  computation. An evidential router produces sparse top-2 expert weights and
  router uncertainty. Four expert outputs are mixed and interpolated with a
  shared fallback by a parameter-free gate formed from routing-confidence,
  probe-uncertainty, and router-uncertainty factors. The resulting adaptive
  output is added to the backbone feature through a positive residual scale.
  Dashed lines denote uncertainty or control signals. Solid lines denote data
  flow.}
  \label{fig:app-ugmoe}
\end{figure}

\bibliographystyle{cas-model2-names}
\bibliography{cas-refs}

\begin{thebibliography}{41}
\expandafter\ifx\csname natexlab\endcsname\relax\def\natexlab#1{#1}\fi
\providecommand{\url}[1]{\texttt{#1}}
\providecommand{\href}[2]{#2}
\providecommand{\path}[1]{#1}
\providecommand{\DOIprefix}{doi:}
\providecommand{\ArXivprefix}{arXiv:}
\providecommand{\URLprefix}{URL: }
\providecommand{\Pubmedprefix}{pmid:}
\providecommand{\doi}[1]{\href{http://dx.doi.org/#1}{\path{#1}}}
\providecommand{\Pubmed}[1]{\href{pmid:#1}{\path{#1}}}
\providecommand{\bibinfo}[2]{#2}
\ifx\xfnm\relax \def\xfnm[#1]{\unskip,\space#1}\fi
\bibitem[{Al-Dhabyani et~al.(2020)Al-Dhabyani, Gomaa, Khaled and Fahmy}]{AlDhabyani2020BUSI}
\bibinfo{author}{Al-Dhabyani, W.}, \bibinfo{author}{Gomaa, M.}, \bibinfo{author}{Khaled, H.}, \bibinfo{author}{Fahmy, A.}, \bibinfo{year}{2020}.
\newblock \bibinfo{title}{Dataset of breast ultrasound images}.
\newblock \bibinfo{journal}{Data in brief} \bibinfo{volume}{28}, \bibinfo{pages}{104863}.
\bibitem[{{amritarajput54}(n.d.)}]{KaggleOvaryUS}
\bibinfo{author}{{amritarajput54}}, \bibinfo{year}{n.d.}
\newblock \bibinfo{title}{Ovarian ultrasound dataset (infected / not infected)}.
\newblock \bibinfo{howpublished}{Kaggle dataset}.
\newblock \URLprefix \url{https://www.kaggle.com/datasets/amritarajput54/hmmmmm}. \bibinfo{note}{public Kaggle release; no formal author or publication year provided on the Kaggle page. Two classes: infected (781) / not infected (1143).}
\bibitem[{Anitha(2024)}]{Anitha2024Fetal}
\bibinfo{author}{Anitha, A.}, \bibinfo{year}{2024}.
\newblock \bibinfo{title}{Ultrasound fetus dataset}.
\newblock \bibinfo{howpublished}{Mendeley Data, V1}.
\newblock \DOIprefix\doi{10.17632/yrzzw9m6kk.1}.
\bibitem[{Born et~al.(2020)Born, Br{\"a}ndle, Cossio, Disdier, Goulet, Roulin and Wiedemann}]{Born2020POCUS}
\bibinfo{author}{Born, J.}, \bibinfo{author}{Br{\"a}ndle, G.}, \bibinfo{author}{Cossio, M.}, \bibinfo{author}{Disdier, M.}, \bibinfo{author}{Goulet, J.}, \bibinfo{author}{Roulin, J.}, \bibinfo{author}{Wiedemann, N.}, \bibinfo{year}{2020}.
\newblock \bibinfo{title}{Pocovid-net: automatic detection of covid-19 from a new lung ultrasound imaging dataset (pocus)}.
\newblock \bibinfo{journal}{arXiv preprint arXiv:2004.12084} .
\bibitem[{Chen et~al.(2025)Chen, Zeng, Wang, Wan, Ning, Liao, Zhang and Chen}]{Chen2025COME}
\bibinfo{author}{Chen, L.}, \bibinfo{author}{Zeng, Y.}, \bibinfo{author}{Wang, Y.}, \bibinfo{author}{Wan, P.}, \bibinfo{author}{Ning, G.}, \bibinfo{author}{Liao, H.}, \bibinfo{author}{Zhang, D.}, \bibinfo{author}{Chen, F.}, \bibinfo{year}{2025}.
\newblock \bibinfo{title}{Come: Dual structure-semantic learning with collaborative moe for universal lesion detection across heterogeneous ultrasound datasets}, in: \bibinfo{booktitle}{Proceedings of the IEEE/CVF International Conference on Computer Vision}, pp. \bibinfo{pages}{21460--21470}.
\bibitem[{Cofta(2025)}]{cofta2025predictability}
\bibinfo{author}{Cofta, P.}, \bibinfo{year}{2025}.
\newblock \bibinfo{title}{The predictability of the effectiveness of chains of classifiers in the out-of-domain detection}.
\newblock \bibinfo{journal}{Engineering Applications of Artificial Intelligence} \bibinfo{volume}{139}, \bibinfo{pages}{109682}.
\bibitem[{El-Assiouti et~al.(2024)El-Assiouti, Hamed, Khattab and Ebied}]{el2024hdkd}
\bibinfo{author}{El-Assiouti, O.S.}, \bibinfo{author}{Hamed, G.}, \bibinfo{author}{Khattab, D.}, \bibinfo{author}{Ebied, H.M.}, \bibinfo{year}{2024}.
\newblock \bibinfo{title}{Hdkd: Hybrid data-efficient knowledge distillation network for medical image classification}.
\newblock \bibinfo{journal}{Engineering Applications of Artificial Intelligence} \bibinfo{volume}{138}, \bibinfo{pages}{109430}.
\bibitem[{Fu et~al.(2025)Fu, Chen, Liu and Yue}]{Fu2025DEDL}
\bibinfo{author}{Fu, W.}, \bibinfo{author}{Chen, Y.}, \bibinfo{author}{Liu, Y.}, \bibinfo{author}{Yue, X.}, \bibinfo{year}{2025}.
\newblock \bibinfo{title}{D-edl: Differential evidential deep learning for robust medical out-of-distribution detection}.
\newblock \bibinfo{journal}{Medical Image Analysis} , \bibinfo{pages}{103888}.
\bibitem[{Geifman et~al.(2018)Geifman, Uziel and El-Yaniv}]{Geifman2018AURC}
\bibinfo{author}{Geifman, Y.}, \bibinfo{author}{Uziel, G.}, \bibinfo{author}{El-Yaniv, R.}, \bibinfo{year}{2018}.
\newblock \bibinfo{title}{Bias-reduced uncertainty estimation for deep neural classifiers}.
\newblock \bibinfo{journal}{arXiv preprint arXiv:1805.08206} .
\bibitem[{Guo et~al.(2017)Guo, Pleiss, Sun and Weinberger}]{Guo2017Calibration}
\bibinfo{author}{Guo, C.}, \bibinfo{author}{Pleiss, G.}, \bibinfo{author}{Sun, Y.}, \bibinfo{author}{Weinberger, K.Q.}, \bibinfo{year}{2017}.
\newblock \bibinfo{title}{On calibration of modern neural networks}, in: \bibinfo{booktitle}{International conference on machine learning}, \bibinfo{organization}{PMLR}. pp. \bibinfo{pages}{1321--1330}.
\bibitem[{Howard et~al.(2019)Howard, Sandler, Chu, Chen, Chen, Tan, Wang, Zhu, Pang, Vasudevan et~al.}]{Howard2019MobileNetV3}
\bibinfo{author}{Howard, A.}, \bibinfo{author}{Sandler, M.}, \bibinfo{author}{Chu, G.}, \bibinfo{author}{Chen, L.C.}, \bibinfo{author}{Chen, B.}, \bibinfo{author}{Tan, M.}, \bibinfo{author}{Wang, W.}, \bibinfo{author}{Zhu, Y.}, \bibinfo{author}{Pang, R.}, \bibinfo{author}{Vasudevan, V.}, et~al., \bibinfo{year}{2019}.
\newblock \bibinfo{title}{Searching for mobilenetv3}, in: \bibinfo{booktitle}{Proceedings of the IEEE/CVF international conference on computer vision}, pp. \bibinfo{pages}{1314--1324}.
\bibitem[{Huang et~al.(2023)Huang, Zhang, Zhang, Li, Ma, Deng, Shen, Wang, Mei and Lei}]{Huang2023BUSIWHU}
\bibinfo{author}{Huang, J.}, \bibinfo{author}{Zhang, J.}, \bibinfo{author}{Zhang, Y.}, \bibinfo{author}{Li, X.}, \bibinfo{author}{Ma, X.}, \bibinfo{author}{Deng, J.}, \bibinfo{author}{Shen, H.}, \bibinfo{author}{Wang, D.}, \bibinfo{author}{Mei, L.}, \bibinfo{author}{Lei, C.}, \bibinfo{year}{2023}.
\newblock \bibinfo{title}{{BUSI\_WHU}: Breast cancer ultrasound image dataset}.
\newblock \bibinfo{journal}{Mendeley Data} \bibinfo{volume}{3}, \bibinfo{pages}{2025}.
\bibitem[{Hung et~al.(2024)Hung, Zheng, Zhao, Pang, Terzopoulos and Sung}]{Hung2024EvidentialProstate}
\bibinfo{author}{Hung, A.L.Y.}, \bibinfo{author}{Zheng, H.}, \bibinfo{author}{Zhao, K.}, \bibinfo{author}{Pang, K.}, \bibinfo{author}{Terzopoulos, D.}, \bibinfo{author}{Sung, K.}, \bibinfo{year}{2024}.
\newblock \bibinfo{title}{Cross-slice attention and evidential critical loss for uncertainty-aware prostate cancer detection}, in: \bibinfo{booktitle}{International Conference on Medical Image Computing and Computer-Assisted Intervention}, \bibinfo{organization}{Springer}. pp. \bibinfo{pages}{113--123}.
\bibitem[{{imtkaggleteam}(n.d.)}]{KaggleKidneyUS}
\bibinfo{author}{{imtkaggleteam}}, \bibinfo{year}{n.d.}
\newblock \bibinfo{title}{Kidney stone classification and object detection (ultrasound dataset)}.
\newblock \bibinfo{howpublished}{Kaggle dataset}.
\newblock \URLprefix \url{https://www.kaggle.com/datasets/imtkaggleteam/kidney-stone-classification-and-object-detection}. \bibinfo{note}{public Kaggle release; no formal author or publication year provided on the Kaggle page. 9416 ultrasound images (4414 normal; 5002 stone), CC BY 4.0.}
\bibitem[{Jacobs et~al.(1991)Jacobs, Jordan, Nowlan and Hinton}]{Jacobs1991MoE}
\bibinfo{author}{Jacobs, R.A.}, \bibinfo{author}{Jordan, M.I.}, \bibinfo{author}{Nowlan, S.J.}, \bibinfo{author}{Hinton, G.E.}, \bibinfo{year}{1991}.
\newblock \bibinfo{title}{Adaptive mixtures of local experts}.
\newblock \bibinfo{journal}{Neural computation} \bibinfo{volume}{3}, \bibinfo{pages}{79--87}.
\bibitem[{Jiang et~al.(2024)Jiang, Zheng, Zhang, Jin, Yuan and Liu}]{Jiang2024MedMoE}
\bibinfo{author}{Jiang, S.}, \bibinfo{author}{Zheng, T.}, \bibinfo{author}{Zhang, Y.}, \bibinfo{author}{Jin, Y.}, \bibinfo{author}{Yuan, L.}, \bibinfo{author}{Liu, Z.}, \bibinfo{year}{2024}.
\newblock \bibinfo{title}{Med-moe: Mixture of domain-specific experts for lightweight medical vision-language models}, in: \bibinfo{booktitle}{Findings of the Association for Computational Linguistics: EMNLP 2024}, pp. \bibinfo{pages}{3843--3860}.
\bibitem[{Jiang et~al.(2025)Jiang, Guo, Xing, Yu, Li, Zhang, Dong and Ta}]{jiang2025prior}
\bibinfo{author}{Jiang, T.}, \bibinfo{author}{Guo, J.}, \bibinfo{author}{Xing, W.}, \bibinfo{author}{Yu, M.}, \bibinfo{author}{Li, Y.}, \bibinfo{author}{Zhang, B.}, \bibinfo{author}{Dong, Y.}, \bibinfo{author}{Ta, D.}, \bibinfo{year}{2025}.
\newblock \bibinfo{title}{A prior segmentation knowledge enhanced deep learning system for the classification of tumors in ultrasound image}.
\newblock \bibinfo{journal}{Engineering Applications of Artificial Intelligence} \bibinfo{volume}{142}, \bibinfo{pages}{109926}.
\bibitem[{Jiao et~al.(2024)Jiao, Zhou, Li, Xia, Huang, Huang, Wang, Zhang, Zhou, Wang et~al.}]{Jiao2024USFM}
\bibinfo{author}{Jiao, J.}, \bibinfo{author}{Zhou, J.}, \bibinfo{author}{Li, X.}, \bibinfo{author}{Xia, M.}, \bibinfo{author}{Huang, Y.}, \bibinfo{author}{Huang, L.}, \bibinfo{author}{Wang, N.}, \bibinfo{author}{Zhang, X.}, \bibinfo{author}{Zhou, S.}, \bibinfo{author}{Wang, Y.}, et~al., \bibinfo{year}{2024}.
\newblock \bibinfo{title}{Usfm: A universal ultrasound foundation model generalized to tasks and organs towards label efficient image analysis}.
\newblock \bibinfo{journal}{Medical Image Analysis} \bibinfo{volume}{96}, \bibinfo{pages}{103202}.
\bibitem[{Jindal and Singh(2025)}]{jindal2025class}
\bibinfo{author}{Jindal, M.}, \bibinfo{author}{Singh, B.}, \bibinfo{year}{2025}.
\newblock \bibinfo{title}{Class imbalance-aware domain specific transfer learning approach for medical image classification: Application on covid-19 detection}.
\newblock \bibinfo{journal}{Engineering Applications of Artificial Intelligence} \bibinfo{volume}{150}, \bibinfo{pages}{110583}.
\bibitem[{Joo et~al.(2023)Joo, Park, Lee, Yoon, Choi and Choi}]{Joo2023Liver}
\bibinfo{author}{Joo, Y.}, \bibinfo{author}{Park, H.C.}, \bibinfo{author}{Lee, O.J.}, \bibinfo{author}{Yoon, C.}, \bibinfo{author}{Choi, M.H.}, \bibinfo{author}{Choi, C.}, \bibinfo{year}{2023}.
\newblock \bibinfo{title}{Classification of liver fibrosis from heterogeneous ultrasound image}.
\newblock \bibinfo{journal}{IEEE Access} \bibinfo{volume}{11}, \bibinfo{pages}{9920--9930}.
\bibitem[{Kang et~al.(2024)Kang, Lao, Gao, Liu, Yi, Ma, Zhang and Li}]{Kang2024UltrasoundMIM}
\bibinfo{author}{Kang, Q.}, \bibinfo{author}{Lao, Q.}, \bibinfo{author}{Gao, J.}, \bibinfo{author}{Liu, J.}, \bibinfo{author}{Yi, H.}, \bibinfo{author}{Ma, B.}, \bibinfo{author}{Zhang, X.}, \bibinfo{author}{Li, K.}, \bibinfo{year}{2024}.
\newblock \bibinfo{title}{Deblurring masked image modeling for ultrasound image analysis}.
\newblock \bibinfo{journal}{Medical Image Analysis} \bibinfo{volume}{97}, \bibinfo{pages}{103256}.
\bibitem[{Karim(2025)}]{Karim2025MSStateUS}
\bibinfo{author}{Karim, S.Z.}, \bibinfo{year}{2025}.
\newblock \bibinfo{title}{Abdominal ultrasound image dataset for organ classification and disease detection}.
\newblock \bibinfo{howpublished}{Mississippi State University Research Data, item 5}.
\newblock \URLprefix \url{https://scholarsjunction.msstate.edu/research-data/5/}. \bibinfo{note}{kidney and pelvic (UB/prostate/uterus) ultrasound subsets used in this work.}
\bibitem[{Krishna and Kokil(2024)}]{krishna2024standard}
\bibinfo{author}{Krishna, T.B.}, \bibinfo{author}{Kokil, P.}, \bibinfo{year}{2024}.
\newblock \bibinfo{title}{Standard fetal ultrasound plane classification based on stacked ensemble of deep learning models}.
\newblock \bibinfo{journal}{Expert Systems with Applications} \bibinfo{volume}{238}, \bibinfo{pages}{122153}.
\bibitem[{Kurz et~al.(2022)Kurz, Hauser, Mehrtens, Krieghoff-Henning, Hekler, Kather, Fr{\"o}hling, Von~Kalle and Brinker}]{Kurz2022UncertaintyMedical}
\bibinfo{author}{Kurz, A.}, \bibinfo{author}{Hauser, K.}, \bibinfo{author}{Mehrtens, H.A.}, \bibinfo{author}{Krieghoff-Henning, E.}, \bibinfo{author}{Hekler, A.}, \bibinfo{author}{Kather, J.N.}, \bibinfo{author}{Fr{\"o}hling, S.}, \bibinfo{author}{Von~Kalle, C.}, \bibinfo{author}{Brinker, T.J.}, \bibinfo{year}{2022}.
\newblock \bibinfo{title}{Uncertainty estimation in medical image classification: systematic review}.
\newblock \bibinfo{journal}{JMIR Medical Informatics} \bibinfo{volume}{10}, \bibinfo{pages}{e36427}.
\bibitem[{Lemay et~al.(2022)Lemay, Hoebel, Bridge, Befano, De~Sanjos{\'e}, Egemen, Rodriguez, Schiffman, Campbell and Kalpathy-Cramer}]{Lemay2022MCDropoutMedical}
\bibinfo{author}{Lemay, A.}, \bibinfo{author}{Hoebel, K.}, \bibinfo{author}{Bridge, C.P.}, \bibinfo{author}{Befano, B.}, \bibinfo{author}{De~Sanjos{\'e}, S.}, \bibinfo{author}{Egemen, D.}, \bibinfo{author}{Rodriguez, A.C.}, \bibinfo{author}{Schiffman, M.}, \bibinfo{author}{Campbell, J.P.}, \bibinfo{author}{Kalpathy-Cramer, J.}, \bibinfo{year}{2022}.
\newblock \bibinfo{title}{Improving the repeatability of deep learning models with monte carlo dropout}.
\newblock \bibinfo{journal}{NPJ digital medicine} \bibinfo{volume}{5}, \bibinfo{pages}{174}.
\bibitem[{Li et~al.(2025)Li, Wu, Gu, Xu, Chen, Cai and Bu}]{Li2025MoESAM}
\bibinfo{author}{Li, R.}, \bibinfo{author}{Wu, L.}, \bibinfo{author}{Gu, J.}, \bibinfo{author}{Xu, Q.}, \bibinfo{author}{Chen, W.}, \bibinfo{author}{Cai, X.}, \bibinfo{author}{Bu, J.}, \bibinfo{year}{2025}.
\newblock \bibinfo{title}{Moe-sam: Enhancing sam for medical image segmentation with mixture-of-experts}, in: \bibinfo{booktitle}{International Conference on Medical Image Computing and Computer-Assisted Intervention}, \bibinfo{organization}{Springer}. pp. \bibinfo{pages}{367--377}.
\bibitem[{Madhu et~al.(2024)Madhu, Kautish, Gupta, Nagachandrika, Biju and Kumar}]{Madhu2024POCUS}
\bibinfo{author}{Madhu, G.}, \bibinfo{author}{Kautish, S.}, \bibinfo{author}{Gupta, Y.}, \bibinfo{author}{Nagachandrika, G.}, \bibinfo{author}{Biju, S.M.}, \bibinfo{author}{Kumar, M.}, \bibinfo{year}{2024}.
\newblock \bibinfo{title}{Xcovnet: An optimized xception convolutional neural network for classification of covid-19 from point-of-care lung ultrasound images}.
\newblock \bibinfo{journal}{Multimedia Tools and Applications} \bibinfo{volume}{83}, \bibinfo{pages}{33653--33674}.
\bibitem[{Ovadia et~al.(2019)Ovadia, Fertig, Ren, Nado, Sculley, Nowozin, Dillon, Lakshminarayanan and Snoek}]{Ovadia2019Trust}
\bibinfo{author}{Ovadia, Y.}, \bibinfo{author}{Fertig, E.}, \bibinfo{author}{Ren, J.}, \bibinfo{author}{Nado, Z.}, \bibinfo{author}{Sculley, D.}, \bibinfo{author}{Nowozin, S.}, \bibinfo{author}{Dillon, J.}, \bibinfo{author}{Lakshminarayanan, B.}, \bibinfo{author}{Snoek, J.}, \bibinfo{year}{2019}.
\newblock \bibinfo{title}{Can you trust your model's uncertainty? evaluating predictive uncertainty under dataset shift}.
\newblock \bibinfo{journal}{Advances in neural information processing systems} \bibinfo{volume}{32}.
\bibitem[{Pang et~al.(2025)Pang, Tang, Song, Qu, Chen, Xiong, Feng and Chen}]{Pang2025Thyroid}
\bibinfo{author}{Pang, Y.Y.}, \bibinfo{author}{Tang, Z.Q.}, \bibinfo{author}{Song, C.}, \bibinfo{author}{Qu, N.}, \bibinfo{author}{Chen, J.Y.}, \bibinfo{author}{Xiong, D.D.}, \bibinfo{author}{Feng, Z.B.}, \bibinfo{author}{Chen, G.}, \bibinfo{year}{2025}.
\newblock \bibinfo{title}{Construction of a multimodal machine learning model for papillary thyroid carcinoma based on pathomics and ultrasound radiomics dataset}.
\newblock \bibinfo{journal}{Data in Brief} \bibinfo{volume}{60}, \bibinfo{pages}{111583}.
\bibitem[{Riquelme et~al.(2021)Riquelme, Puigcerver, Mustafa, Neumann, Jenatton, Susano~Pinto, Keysers and Houlsby}]{Riquelme2021VMoE}
\bibinfo{author}{Riquelme, C.}, \bibinfo{author}{Puigcerver, J.}, \bibinfo{author}{Mustafa, B.}, \bibinfo{author}{Neumann, M.}, \bibinfo{author}{Jenatton, R.}, \bibinfo{author}{Susano~Pinto, A.}, \bibinfo{author}{Keysers, D.}, \bibinfo{author}{Houlsby, N.}, \bibinfo{year}{2021}.
\newblock \bibinfo{title}{Scaling vision with sparse mixture of experts}.
\newblock \bibinfo{journal}{Advances in Neural Information Processing Systems} \bibinfo{volume}{34}, \bibinfo{pages}{8583--8595}.
\bibitem[{Saibene et~al.(2021)Saibene, Assale and Giltri}]{saibene2021expert}
\bibinfo{author}{Saibene, A.}, \bibinfo{author}{Assale, M.}, \bibinfo{author}{Giltri, M.}, \bibinfo{year}{2021}.
\newblock \bibinfo{title}{Expert systems: Definitions, advantages and issues in medical field applications}.
\newblock \bibinfo{journal}{Expert Systems with Applications} \bibinfo{volume}{177}, \bibinfo{pages}{114900}.
\bibitem[{Sensoy et~al.(2018)Sensoy, Kaplan and Kandemir}]{Sensoy2018EDL}
\bibinfo{author}{Sensoy, M.}, \bibinfo{author}{Kaplan, L.}, \bibinfo{author}{Kandemir, M.}, \bibinfo{year}{2018}.
\newblock \bibinfo{title}{Evidential deep learning to quantify classification uncertainty}.
\newblock \bibinfo{journal}{Advances in neural information processing systems} \bibinfo{volume}{31}.
\bibitem[{{\c{S}}ensoy et~al.(2025){\c{S}}ensoy, Kaplan, Julier, Saleki and Cerutti}]{csensoy2025risk}
\bibinfo{author}{{\c{S}}ensoy, M.}, \bibinfo{author}{Kaplan, L.M.}, \bibinfo{author}{Julier, S.}, \bibinfo{author}{Saleki, M.}, \bibinfo{author}{Cerutti, F.}, \bibinfo{year}{2025}.
\newblock \bibinfo{title}{Risk-aware classification via uncertainty quantification}.
\newblock \bibinfo{journal}{Expert Systems with Applications} \bibinfo{volume}{265}, \bibinfo{pages}{125906}.
\bibitem[{Shaddock and Smith(2022)}]{Shaddock2022PortablePOCUS}
\bibinfo{author}{Shaddock, L.}, \bibinfo{author}{Smith, T.}, \bibinfo{year}{2022}.
\newblock \bibinfo{title}{Potential for use of portable ultrasound devices in rural and remote settings in australia and other developed countries: a systematic review}.
\newblock \bibinfo{journal}{Journal of Multidisciplinary Healthcare} , \bibinfo{pages}{605--625}.
\bibitem[{Shazeer et~al.(2017)Shazeer, Mirhoseini, Maziarz, Davis, Le, Hinton and Dean}]{Shazeer2017OutrageouslyMoE}
\bibinfo{author}{Shazeer, N.}, \bibinfo{author}{Mirhoseini, A.}, \bibinfo{author}{Maziarz, K.}, \bibinfo{author}{Davis, A.}, \bibinfo{author}{Le, Q.}, \bibinfo{author}{Hinton, G.}, \bibinfo{author}{Dean, J.}, \bibinfo{year}{2017}.
\newblock \bibinfo{title}{Outrageously large neural networks: The sparsely-gated mixture-of-experts layer}.
\newblock \bibinfo{journal}{arXiv preprint arXiv:1701.06538} .
\bibitem[{Wang et~al.(2025a)Wang, Guo, Chen, Guo, He, Chen and Zhang}]{wang2025abus}
\bibinfo{author}{Wang, C.}, \bibinfo{author}{Guo, Y.}, \bibinfo{author}{Chen, H.}, \bibinfo{author}{Guo, Q.}, \bibinfo{author}{He, H.}, \bibinfo{author}{Chen, L.}, \bibinfo{author}{Zhang, Q.}, \bibinfo{year}{2025}a.
\newblock \bibinfo{title}{Abus-net: Graph convolutional network with multi-scale features for breast cancer diagnosis using automated breast ultrasound}.
\newblock \bibinfo{journal}{Expert Systems with Applications} \bibinfo{volume}{273}, \bibinfo{pages}{126978}.
\bibitem[{Wang et~al.(2025b)Wang, Nie, Duan, Zhao and Chen}]{wang2025mixture}
\bibinfo{author}{Wang, J.}, \bibinfo{author}{Nie, L.}, \bibinfo{author}{Duan, J.}, \bibinfo{author}{Zhao, H.}, \bibinfo{author}{Chen, C.P.}, \bibinfo{year}{2025}b.
\newblock \bibinfo{title}{Mixture-of-experts-based broad learning system and its applications}.
\newblock \bibinfo{journal}{Expert Systems with Applications} \bibinfo{volume}{269}, \bibinfo{pages}{126389}.
\bibitem[{Woo et~al.(2018)Woo, Park, Lee and Kweon}]{Woo2018CBAM}
\bibinfo{author}{Woo, S.}, \bibinfo{author}{Park, J.}, \bibinfo{author}{Lee, J.Y.}, \bibinfo{author}{Kweon, I.S.}, \bibinfo{year}{2018}.
\newblock \bibinfo{title}{Cbam: Convolutional block attention module}, in: \bibinfo{booktitle}{Proceedings of the European conference on computer vision (ECCV)}, pp. \bibinfo{pages}{3--19}.
\bibitem[{Xia et~al.(2024)Xia, Dang, Han, Qendro and Mascolo}]{Xia2024ClassBalancedEDL}
\bibinfo{author}{Xia, T.}, \bibinfo{author}{Dang, T.}, \bibinfo{author}{Han, J.}, \bibinfo{author}{Qendro, L.}, \bibinfo{author}{Mascolo, C.}, \bibinfo{year}{2024}.
\newblock \bibinfo{title}{Uncertainty-aware health diagnostics via class-balanced evidential deep learning}.
\newblock \bibinfo{journal}{IEEE Journal of Biomedical and Health Informatics} \bibinfo{volume}{28}, \bibinfo{pages}{6417--6428}.
\bibitem[{Yang and Yee(2024)}]{yang2024towards}
\bibinfo{author}{Yang, S.}, \bibinfo{author}{Yee, K.}, \bibinfo{year}{2024}.
\newblock \bibinfo{title}{Towards reliable uncertainty quantification via deep ensemble in multi-output regression task}.
\newblock \bibinfo{journal}{Engineering Applications of Artificial Intelligence} \bibinfo{volume}{132}, \bibinfo{pages}{107871}.
\bibitem[{Zhu et~al.(2025)Zhu, Li, Xu and Song}]{zhu2025lightweight}
\bibinfo{author}{Zhu, Y.}, \bibinfo{author}{Li, L.}, \bibinfo{author}{Xu, X.}, \bibinfo{author}{Song, Y.}, \bibinfo{year}{2025}.
\newblock \bibinfo{title}{Lightweight multi-modal ultrasound classification with plug-and-play mcsa and cmufa modules: Achieving high accuracy with low complexity}, in: \bibinfo{booktitle}{2025 IEEE 15th International Conference on Signal Processing, Communications and Computing (ICSPCC)}, \bibinfo{organization}{IEEE}. pp. \bibinfo{pages}{1--6}.

\end{thebibliography}

\end{document}